\documentclass{article}

\usepackage[nonatbib, preprint]{neurips_data_2024}

\usepackage[utf8]{inputenc} 
\usepackage[T1]{fontenc}    
\usepackage{hyperref}       
\usepackage{url}            
\usepackage{booktabs}       
\usepackage{amsfonts}       
\usepackage{nicefrac}       
\usepackage{microtype}      
\usepackage{xcolor}         
\usepackage{graphicx}
\usepackage{subcaption}
\usepackage{multirow}
\usepackage{cleveref}
\usepackage{float}
\usepackage{comment}

\definecolor{vlm_green}{RGB}{0, 158, 115}
\definecolor{vlm_orange}{RGB}{230, 159, 0}
\definecolor{vlm_red}{RGB}{213, 94, 0}

\title{CableInspect-AD: An Expert-Annotated Anomaly Detection Dataset}

\author{%
Akshatha Arodi$^{1*}$ \quad 
Margaux Luck$^{1*}$ \quad 
Jean-Luc Bedwani$^2$ \\ 
\textbf{Aldo Zaimi}$^1$  \quad 
\textbf{Ge Li}$^1$ \quad 
\textbf{Nicolas Pouliot}$^2$ \quad 
\textbf{Julien Beaudry}$^2$ \quad 
\textbf{Ga\'etan Marceau Caron}$^1$\\
$^1$Mila - Quebec AI Institute \quad $^2$IREQ - Institut de recherche d'Hydro-Québec\\
$^{*}$equal contributions\\
}

\begin{document}

\maketitle

\begin{abstract}
  Machine learning models are increasingly being deployed in real-world contexts. However, systematic studies on their transferability to specific and critical applications are underrepresented in the research literature. An important example is visual anomaly detection (VAD) for robotic power line inspection. While existing VAD methods perform well in controlled environments, real-world scenarios present diverse and unexpected anomalies that current datasets fail to capture. To address this gap, we introduce \textit{CableInspect-AD}, a high-quality, publicly available dataset created and annotated by domain experts from Hydro-Québec, a Canadian public utility. This dataset includes high-resolution images with challenging real-world anomalies, covering defects with varying severity levels. 
  To address the challenges of collecting diverse anomalous and nominal examples for setting a detection threshold, we propose an enhancement to the celebrated PatchCore algorithm. 
  This enhancement enables its use in scenarios with limited labeled data. We also present a comprehensive evaluation protocol based on cross-validation to assess models' performances. We evaluate our \textit{Enhanced-PatchCore} for few-shot and many-shot detection, and Vision-Language Models for zero-shot detection. While promising, these models struggle to detect all anomalies, highlighting the dataset's value as a challenging benchmark for the broader research community. Project page: \url{https://mila-iqia.github.io/cableinspect-ad/}.
\end{abstract}

\section{Introduction}

\begin{figure}[h!]
  \centering
  \includegraphics[width=\textwidth,trim={0 10px 0 10px},clip]{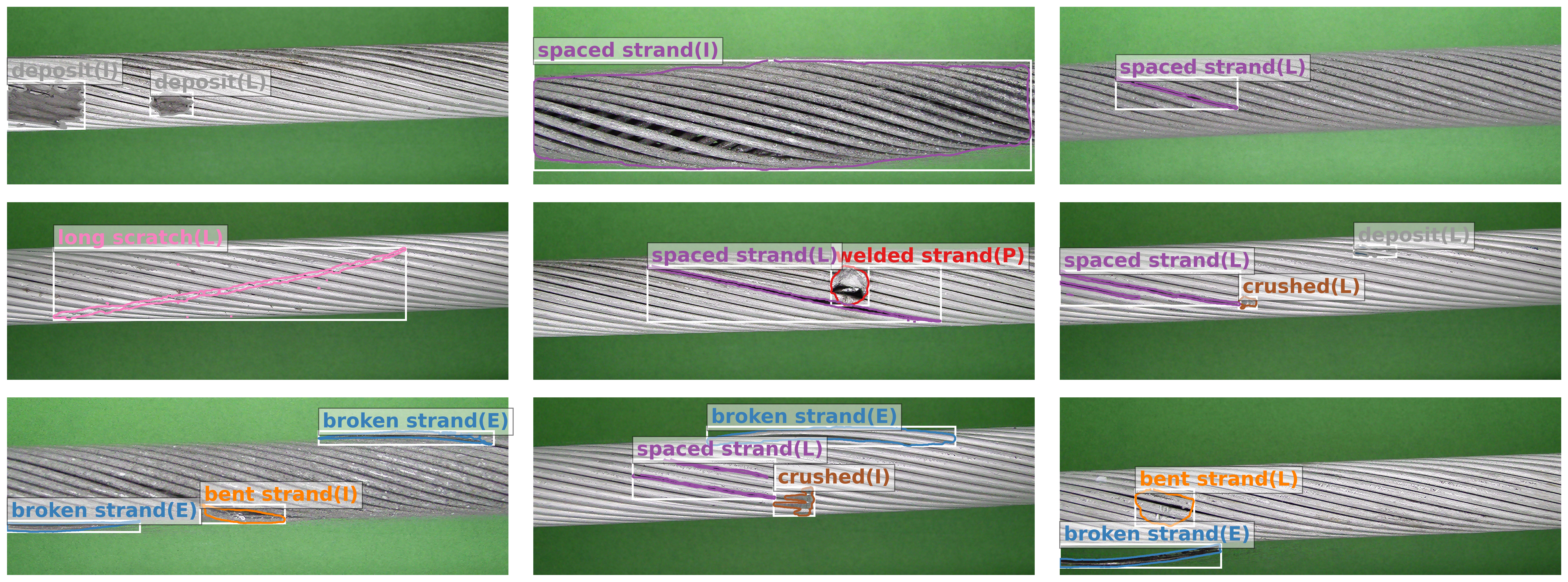}
  \caption{Examples of anomalies. On each image, the anomaly types (grades) are annotated (masks outlined). The grades here are (I)mportant, (L)ight, (C)omplete, (E)xtracted and (D)eep. Anomalies such as long scratches(I) are hard to spot, whereas deposit(I) and spaced strands(I) are easier.}
  \label{fig:anomalies_mosaic}
\end{figure}

Machine learning is increasingly applied across diverse industrial fields such as robotics, genomics, climate and materials science due to the impressive performance of large pre-trained models. As the community looks towards deploying these models in specialized domains where their effectiveness remains uncertain, there is a pressing need to improve their transferability in these contexts. This underscores the necessity for tailored datasets by domain experts. Visual anomaly detection (VAD) in a specific industrial context, exemplifies a critical application, promising cost reduction, time savings, and enhanced safety measures by enabling preventive maintenance. While existing VAD methods perform well in controlled environments, real-world scenarios present diverse and unexpected anomalies that current datasets fail to capture. Public VAD datasets, such as MvTec AD~\cite{bergmann2019mvtec}, VisA~\cite{zou2022spot}, and MVTec LOCO AD~\cite{bergmann2022beyond}, focus mainly on objects and textures in a controlled manufacturing context, thus limiting the scope of potential anomalies. Moreover, these datasets do not account for scenarios with significant variations of the same object, further complicating AD in real-world applications. For instance, objects may exhibit substantial differences when viewed indoors versus outdoors due to varying operational conditions and environmental factors such as lighting and weather. Additionally, wear and tear over time can introduce anomalies that evolve, creating multiple views and states of the same object. Compounding the complexity, images may contain more than one anomaly, requiring models to discern and identify multiple issues simultaneously.

Robotic power line inspection represents a specialized and highly challenging domain characterized by a wide range of anomalies, further complicated by the changing appearance of cables due to natural wear. Recognizing the importance of open-science and transparency in evaluating machine learning models for such complex real-world applications, there is a clear need for more public industrial datasets. To this end, we introduce \textit{CableInspect-AD} (see \Cref{fig:anomalies_mosaic}), a high-quality, publicly available dataset created and annotated by domain experts from Hydro-Québec\footnote{\url{https://en.wikipedia.org/wiki/Hydro-Qu\%C3\%A9bec}}, a Canadian public utility. It features 4,798 high-resolution images and 6,023 annotated anomalies across three types of power line cables. These anomalies represent the seven most common defect types listed by Hydro-Québec, with varying severity levels. They are meticulously crafted by experts and are annotated at the image level, the pixel-level, and with bounding boxes, to provide a detailed categorization of those anomalies both by type and by severity level.

To address the challenges of collecting diverse anomalous and nominal examples for setting a detection threshold, we introduce \textit{Enhanced-PatchCore}, an improved approach that sets thresholds using only a training set with a few nominal images. This approach enhances adaptability and robustness to diverse anomaly types encountered in real-world industrial settings. Furthermore, this enhancement enables its application in scenarios with limited labeled data. We define a comprehensive evaluation protocol based on cross-validation and evaluate \textit{Enhanced-PatchCore} for few-shot and many-shot detection. To further eliminate the need for a train set, we seek to use open\footnote{Open models here are defined as those with widely accessible weights.} conversational Vision-Language Models (VLMs)~\cite{wang2023cogvlm,liu2023improved}, which have demonstrated impressive capacity in zero-shot settings on tasks like Visual Question Answering (VQA) and image captioning. To the best of our knowledge, this is among the first attempts to utilize open conversational VLMs for zero-shot VAD in this context. Our findings indicate that the baselines show promising results in detecting anomalies on the cables. However, they struggle with certain types and grades of anomalies, highlighting the need for further research in real-world industrial contexts. By introducing \textit{CableInspect-AD}, we aim to push the frontiers of VAD and demonstrate its potential to generalize to complex, real-world domains.

\section{Related work}
\label{sec:related work}

\paragraph{Datasets} Industrial VAD datasets such as MvTec AD \cite{bergmann2019mvtec}, VisA \cite{zou2022spot} and MVTec LOCO AD \cite{bergmann2022beyond} are commonly used to evaluate VAD methods, focusing primarily on objects and textures inspection in a controlled environment. However, despite their popularity, the evaluation results from these datasets may not translate to specialized domains, such as VAD in power line cables. Specifically, VisA excludes cables, and while MVTec AD does feature a cable category, it focuses on the cross-sectional aspect of cables and lacks the nuanced defects found in power line cables.

Existing public power line inspection datasets predominantly focus on specific power line components like transmission towers and insulators~\cite{prates2019insulator,tao2018detection,tomaszewski2018collection, abdelfattah2020ttpla,vieira2021stn,de2022ptl,bian2019monocular} and often overlook intricacies and anomalies on cables. While the InsPLAD dataset~\cite{vieira2023insplad} addresses both object detection (InsPLAD-det) and VAD (InsPLAD-fault), it does not feature anomalies on cables. In contrast, some datasets focus solely on power line cables but are primarily intended for cable detection or segmentation rather than inspection~\cite{cano2023vepl,choi2019real,lee2017weakly,madaan2017wire,Recognizance-2,yetgin2017ground}. For example, datasets designed for aircraft safety~\cite{yetgin2017ground} or autonomous flying vehicles~\cite{choi2019real,madaan2017wire} offer low-resolution, birds-eye-view shots of cables without annotations for cable anomalies. This lack of specialized datasets tailored to power line cable anomalies underscores the need for a new dataset.

\paragraph{Anomaly detection algorithms} VAD in industrial settings predominantly relies on \emph{unsupervised} methods~\cite{liu2023deep}. This preference mainly stems from the ease of obtaining nominal examples compared to the expensive and complex task of specifying expected defect variations.

Consequently, the training set often contains only nominal samples, while the validation and test sets include both anomalous and nominal samples for model evaluation. The taxonomy proposed by~\cite{liu2023deep} classifies these methods into two broad categories: \emph{reconstruction-based} and \emph{feature-embedding-based} approaches.

Reconstruction-based approaches~\cite{zavrtanik2021draem,akcay2019ganomaly,lee2022anovit,shi2016uniad,yang2020dfr,song2021anoseg,you2022adtr,wyatt2022anoddpm} typically involve training encoder-decoder models. During testing, they predict anomalies by comparing the input image with its reconstruction, assuming models will generate errors for anomalies not part of the training set.

Feature-embedding-based approaches ~\cite{wang2021student,li2021cutpaste,rudolph2022fully,tsai2022multi,roth2022towards} on the other hand, employ pre-trained models to generate embeddings for VAD. Among several methods, instance-based approaches are the most effective \cite{liu2023deep}. These methods store normal feature embeddings in a \emph{memory bank}, where embeddings far from those in the memory bank are likely anomalous. Notably, the PatchCore~\cite{roth2022towards} algorithm demonstrates significant advancements, achieving state-of-the-art results on benchmarks such as MVTec AD and VisA, showing promising performance in both few-shot and many-shot settings~\cite{santos2023optimizing}. While methods like PatchCore can work with few nominal examples, they still need a comprehensive set of both nominal and anomalous images to select a threshold, which is impractical in real-world applications where collecting diverse anomalies is difficult. Consequently, these methods often face challenges in generalization, particularly when anomalies are rare and the nominal images are diverse, leading to unreliable performance. 

More recent research has explored the application of large models and VLMs to VAD. Models based on CLIP~\cite{deng2023anovl, jeong2023winclip, chen2023zero}, SAM and GroundingDINO ~\cite{cao2023segment, kirillov2023segment}, and conversational VLMs ~\cite{openai2023gpt4v, cao2023towards, zhang2023exploring, gu2023anomalygpt} have shown promising results. These models leverage the capabilities of VLMs in zero-/few-shot inference and image understanding tasks. For instance, \cite{cao2023towards,zhang2023exploring} demonstrates the potential of GPT-4V's generic capacity on zero-shot VAD tasks. However, it leverages a proprietary model with limited API access. In addition, AnomalyGPT \cite{gu2023anomalygpt}, a conversational VLM fine-tuned for VAD tasks, requires finetuning on a set of nominal and simulated anomalous images, which can be costly and impractical in real-world VAD applications. In contrast, our study explores the use of open conversational VLMs for zero-shot VAD to ensure our comparisons are accessible and replicable within the research community. 

Other recent works on utilizing large models for VAD tasks such as MuSc~\cite
{li2024musc} and APRIL-GAN~\cite{chen2023april}, while demonstrating competitive performances in zero-/few-shot scenarios, do not align with the approach opted in this work. Although MuSc is claimed to be a zero-shot method, it still relies on prior knowledge from a test set, a requirement that is impractical for real-world applications like power line inspection. In addition, the method assumes the test set contains abundant information on both normal and abnormal cues, which is not applicable in settings where only nominal images are available. APRIL-GAN, while achieving good results in certain contexts, requires an additional training phase, which is resource-intensive for training and evaluation on our dataset. Furthermore, WinCLIP~\cite{jeong2023winclip} either matches or outperforms APRIL-GAN in similar contexts, making it a more suitable candidate for initial benchmarking.

\section{CableInspect-AD dataset}

Advances in robotics, exemplified by Hydro-Québec's LineRanger robot~\cite{richard2019lineranger}, have transformed power line inspections, introducing automation for increased efficiency~\cite{alhassan2020power,hamelin2019discrete,richard2019lineranger}. Our \textit{CableInspect-AD} dataset, developed by Hydro-Québec experts, plays a crucial role in furthering robotics through deep learning and serves as a benchmark for developing and evaluating new VAD algorithms with real-world data. It addresses the challenge of detecting rare multi-scale anomalies on power line cables, which vary in wear, color, texture, and braiding. It also facilitates the extension of these techniques to other infrastructure-monitoring areas, such as railways and pipelines, fostering the evaluation of VAD models and the creation of predictive maintenance systems to advance VAD technologies across various sectors.

\paragraph{Creation and annotation} The creation and annotation of \textit{CableInspect-AD} is highly challenging and requires domain expertise. To achieve this, experts selected three cables used in the field. The cables are suspended for image acquisition, and an apparatus with a moving camera is used to capture the images to ensure a uniform background and mimic real-world robotic scenarios. The uniform background was intentionally chosen to minimize distractions and external factors, allowing models to focus solely on detecting anomalies within the object, a practice commonly seen in other VAD benchmarks. Importantly, capturing images while the apparatus is in motion introduces slight disturbances, making the images less perfect compared to datasets like MVTec AD, thereby adding to the dataset's uniqueness and realism. To maximize the use of each cable, both sides (referred to as sides A and B) are utilized.

For each cable side, three videos are recorded at a frame rate of 30 frames per second, consisting of RGBA images at a resolution of 1920$\times$1080 pixels. A total of 18 videos are captured by manually moving a camera along the cables at different speeds, slow enough to capture a defect in several frames. Each pass includes minor rotational variations, up to 20 degrees, and can be taken forward or backward, slightly changing the perspective. The videos are then processed to keep one frame out of three for anomaly annotations, reducing the frame rate to 10 frames per second.

Annotations include image-level labels and bounding boxes, assigned based on expert assessment of the anomaly's appearance in the image. Additionally, per-pixel labels for the first recorded video on each cable are obtained using SAM~\cite{kirillov2023segment} prompted with the bounding boxes and then manually corrected. Depending on the point of view, a defect can be associated with different grades. An image containing at least one bounding box is considered anomalous. Examples of anomalies are shown in \Cref{fig:anomalies_mosaic}, illustrating their varying appearance and complexity. The dataset was annotated by at least four experts who first developed and agreed on guidelines to establish a clear annotation framework. The process was repeated five times until an agreement was achieved. The acquisition process, annotation guide, and details on the annotation process are in Supplementary Material.

\begin{figure}[h!]
   \centering
   \includegraphics[width=0.95\textwidth,trim={0 5px 0 5px},clip]{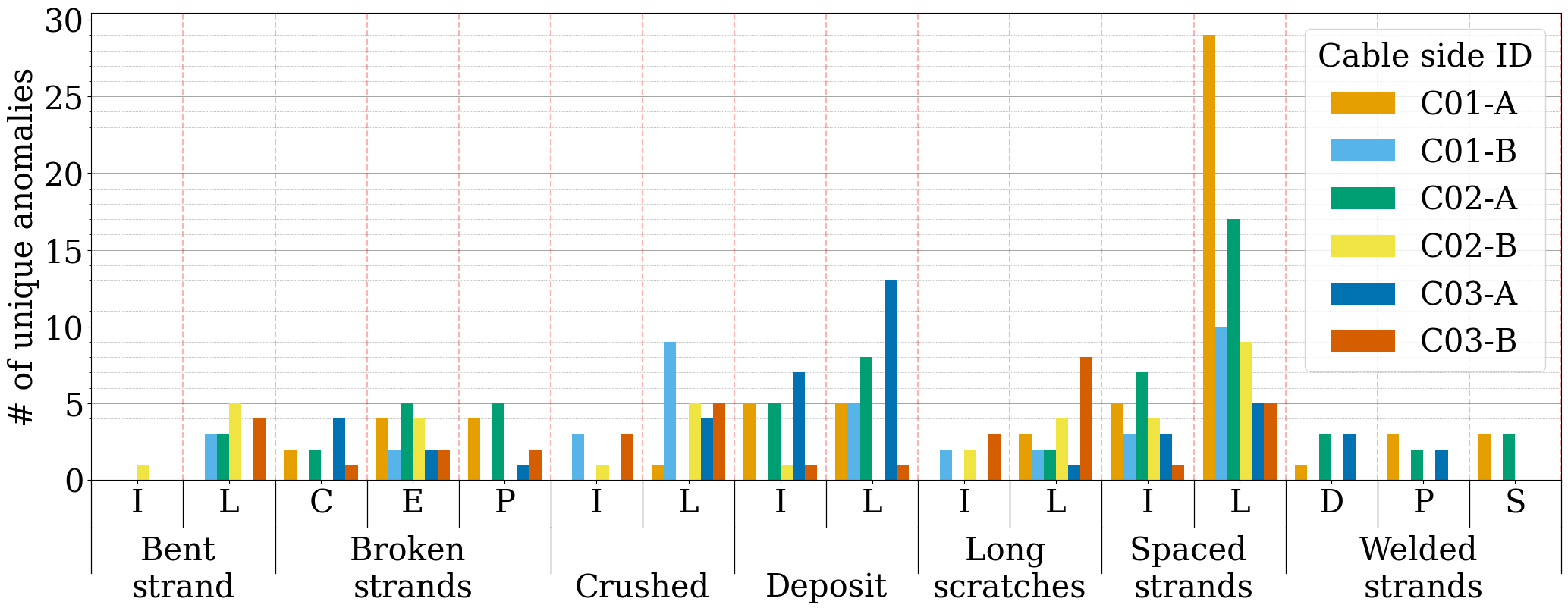}
   \caption{Anomaly types and grades per cable. The grades are (I)mportant, (L)ight, (C)omplete, (E)xtracted, (P)artial, (D)eep and (S)uperficial. The anomalies are not distributed uniformly across all the cables.}
   \label{fig:anomalies_types_grades}
\end{figure}

\paragraph{Statistics} The dataset contains 4,798 annotated images (2,639 anomalous and 2,159 nominal). Among the anomalous images, there are 193 unique anomalies, comprising 110 manually created and 83 pre-existing real-world anomalies. The total number of anomalies annotated is 6,023. The distribution of defects among the three cables is shown in \Cref{fig:anomalies_types_grades}.

\paragraph{Evaluation protocol} To estimate variance in model performance, we use a k-fold cross-validation strategy tailored to our dataset. This approach addresses the high anomaly ratio resulting from the deliberate creation of diverse anomalies, the non-uniform distribution of anomalies, and possible data leakage due to overlapping video frames. Specifically, we split the power line cable dataset into train and test sets using a k-fold sampling strategy based on defect identifiers. For each fold, defect identifiers are randomly selected, and 100 subsequent nominal images are selected for training while preventing overlap between training and test sets using buffers. This process is repeated k times, ensuring a consistent training size but varying test images and anomaly ratios across folds as shown in \Cref{fig:num_test_raw}. More details can be found in the Supplementary Material.

\begin{figure}[h!]
  \centering
  \begin{subfigure}[b]{0.4\textwidth}
    \centering
    \includegraphics[width=\textwidth,trim={0 30px 0 60px},clip]{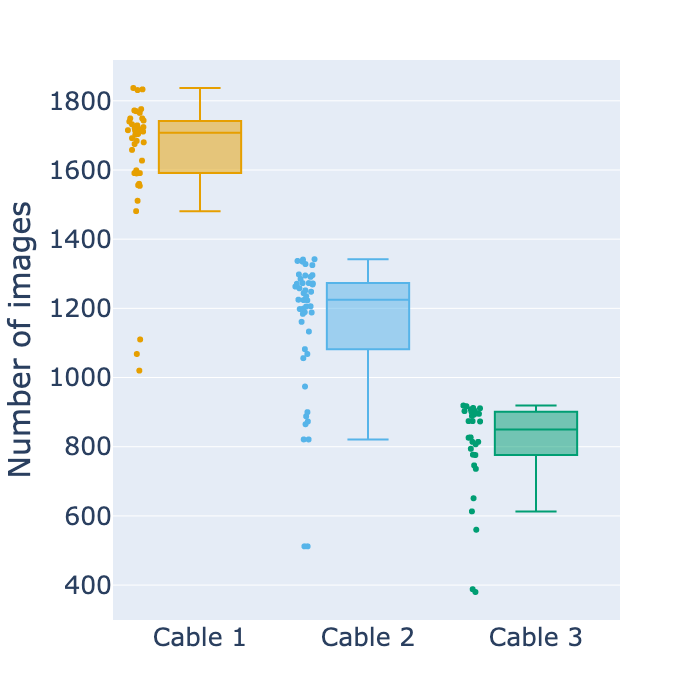}
    \caption{}
  \end{subfigure} 
  \hspace{15px}
  \begin{subfigure}[b]{0.4\textwidth}
    \centering
    \includegraphics[width=\textwidth,trim={0 10px 0 60px},clip]{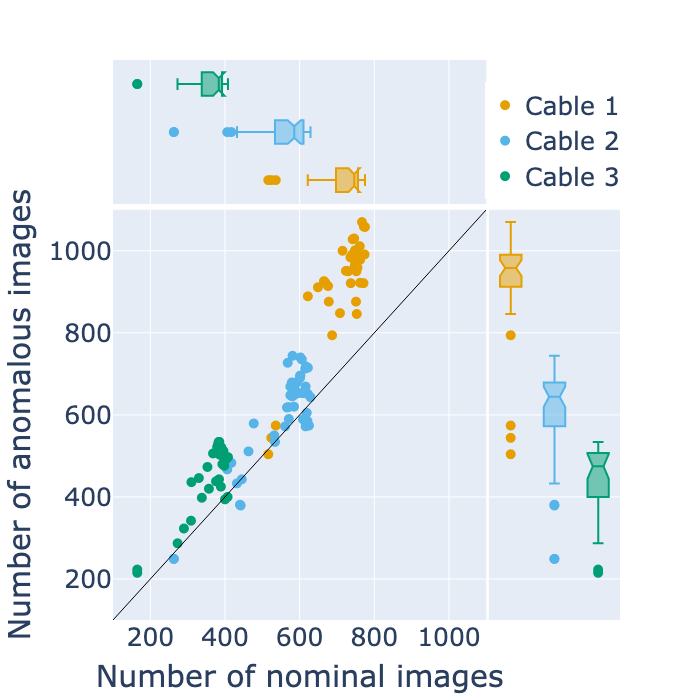}
    \caption{}
  \end{subfigure}
  \caption{The three cables have different numbers of images with varying anomaly ratios in the test set. The cables have 40, 46, and 30 folds, respectively. 
  (a) shows the number of images in the test set over all the folds for each cable (x-axis), and (b) shows the ratio in the test set of the cables. Each point corresponds to the anomaly ratio in a fold. The identity line shows where a balanced dataset would be.}
  \label{fig:num_test_raw}
\end{figure}

\section{Enhanced-PatchCore}

\textit{Enhanced-PatchCore}, built on PatchCore~\cite{roth2022towards}, is an instance-based approach that stores feature embeddings of nominal images in a \emph{memory bank} $\mathcal{M}$ to establish a context during training. This memory bank is then coreset-subsampled~\cite{agarwal2005geometric} to reduce its size.

At test time, the abnormality of a test image $X$ is determined by measuring its distance to the nearest neighbor in the memory bank within the embedding space. This distance, referred to as anomaly score, is defined as:

\begin{equation}
\label{eq:anomaly_score_test}
    S(X) := \max_{e \in \mathcal{P}(X)} d(e, \mathcal{M}) = \max_{e \in \mathcal{P}(X)} \min_{e' \in \mathcal{M}} d(e, e')
\end{equation}

where $\mathcal{P}(X)$ is the set of patch embeddings generated by an image encoder and $d$ is the Euclidean distance.

To decide if an image contains an anomaly from this score, a threshold must be set using a validation set. However, creating a robust validation set with a diverse range of anomalies is prohibitively expensive. Many VAD methods overlook this crucial aspect, either manually setting thresholds or reporting the best F1 score. This is impractical in real-world applications, where thresholds must be carefully calibrated to specific operational requirements and constraints. Therefore, we introduce \textit{Enhanced-PatchCore}, which addresses this challenge by setting a threshold using only the train set. Specifically, it computes anomaly scores of images within the memory bank to estimate the empirical distribution of scores of nominal images. The score $S(X)$ is calculated as follows:

\begin{equation}
\label{eq:anomaly_score}
    \hat{S}(X) := \max_{e \in \mathcal{P}(X)} \min_{e' \in \mathcal{M} \setminus \mathcal{P}(X)} d(e, e')
\end{equation}

Similarly, a segmentation map can be computed by realigning the patch anomaly scores to match the original input resolution by upscaling the scores using bi-linear interpolation. Specifically, the anomaly score at the pixel level for a pixel at coordinates $(i, j)$ in the image, with embedding $e_{i, j}$ is computed using the following equation:

\begin{equation}
    \hat{S}(X_{i,j}) := \min_{e' \in \mathcal{M} \setminus \mathcal{P}(X)} d(e_{i, j}, e'),
\end{equation}

Experimentally, the distribution of $\hat{S}(X)$ closely matches the one from a validation set. We evaluate four thresholding strategies on this estimated empirical distribution: \textit{max}, outliers from a boxplot (\textit{whisker}), percentile estimation from parametric distribution at $95th$ percentile (\textit{beta-prime-$95$}), and percentile estimation from empirical distribution at $95th$ percentile (\textit{empirical-$95$}). Additional details can be found in Supplementary Material.

\section{Experimental setting}
\label{sec:experimental_setting}

Our experimental setup assumes the unavailability of a validation set, reflecting real-world challenges. Furthermore, many VAD methods assume that the training data contains only nominal images, but the presence of contaminated training data with anomalies can significantly reduce performance \cite{yoon2021self}. 

Given the difficulty of avoiding such contamination in specialized domains due to annotation challenges, our setup transitions from many-shot to few-shot and finally to zero-shot settings by gradually reducing the number of examples in the training set until it is completely removed.

To adhere to our setup constraints, we employed pre-trained models without fine-tuning that operate effectively in low-data regimes as baselines. Specifically, we propose \textit{Enhanced-Patchcore} for few-shot and many-shot settings. For the zero-shot setting, we use conversational VLMs including LLaVA 1.5-7B/13B and BakLLaVA-7B,  \cite{liu2023improved}, CogVLM-17B and CogVLM2-19B \cite{wang2023cogvlm}, and a VLM tailored for VAD, WinCLIP~\cite{jeong2023winclip}. The prompt used to get VLMs' predictions is ``\textit{Is there any anomaly or defect in the image. Please answer by Yes or No.}''. For WinCLIP, we use ``cable'' as the object to fill the templates. For the many-shot and few-shot tasks, $N$ images were randomly sampled from the training sections within the k-fold cross-validation. For the zero-shot task, the training sections were entirely discarded. The test sections remain constant within the k-fold across all tasks.

To evaluate our models' performance, we consider threshold-independent metrics Area Under the Precision-Recall curve (AUPR) and Area Under the Receiver Operating Characteristic Curve (AUROC), and threshold-dependent metrics: precision, recall, false positive rate (FPR), false negative rate (FNR) and F1-score at the image level. To compute AUROC and AUPR for conversational VLMs, we adapt the VQAScore \cite{lin2024evaluating} to obtain anomaly scores. Specifically, VQAScore computes the probability of the output token ``\textit{Yes}'' when prompting VLMs with the fixed template ``\textit{Does this figure show [caption]? Please answer yes or no.}''. We use ``\textit{an anomalous or defective cable}'' as ``\textit{[caption]}''. For per-pixel evaluation we use AUPRO~\cite{bergmann2019mvtec}. Additional implementation details are in Supplementary Material.

\section{Results and discussion}
\label{sec:results_and_discussion}

\begin{table}[h!]
    \centering
    \caption{Performance metrics at image-level. Mean and standard deviation are calculated across all cables after averaging over all folds. VLMs and WinCLIP are evaluated in a zero-shot setting, while \textit{Enhanced-PatchCore} is evaluated in a 100-shot setting using the \textit{beta-prime-$95$} thresholding strategy. Thresholded-metrics are not reported for WinCLIP since it necessitates a validation set.}
    \resizebox{\textwidth}{!}{%
    \tiny
    \begin{tabular}{lccccc}
        \toprule
        \textbf{Model}
         &  \textbf{F1 Score} $\uparrow$ & \textbf{FPR} $\downarrow$ & \textbf{AUPR} $\uparrow$ & \textbf{AUROC} $\uparrow$ \\
        \midrule
        LLaVA 1.5-7B & 0.59 $\pm$ 0.07 & 0.32 $\pm$ 0.19 &0.75 $\pm$ 0.05 &0.68 $\pm$ 0.04 \\
        LLaVA 1.5-13B & 0.69 $\pm$ 0.02 & 0.66 $\pm$ 0.21 &0.74 $\pm$ 0.04 & 0.66 $\pm$ 0.03\\
        BakLLaVA-7B & 0.69 $\pm$ 0.02 & 0.53 $\pm$ 0.19  & 0.77 $\pm$ 0.04 &0.71 $\pm$ 0.03\\
        CogVLM-17B& \textbf{0.77 $\pm$ 0.02} & 0.34 $\pm$ 0.21  &0.83 $\pm$ 0.03 & 0.79 $\pm$ 0.04\\
        CogVLM2-19B& 0.66 $\pm$ 0.04 & \textbf{0.04 $\pm$ 0.01}  &\textbf{0.91 $\pm$ 0.02} &\textbf{0.86 $\pm$ 0.03}\\
        WinCLIP & - & - & 0.76 $\pm$ 0.06 & 0.70 $\pm$ 0.04  \\
        \midrule
        \textit{Enhanced-PatchCore} & 0.75 $\pm$ 0.03 & 0.55 $\pm$ 0.19  & 0.84 $\pm$ 0.06 & 0.78 $\pm$ 0.05\\
        \bottomrule
    \end{tabular}}
    \label{tab:performance_metrics}
\end{table}

Table \ref{tab:performance_metrics} summarizes the overall performance of the baseline models and \textit{Enhanced-PatchCore} on our \textit{CableInspect-AD} dataset at image-level. First, we can observe that CogVLM-17B has the best F1 Score, whereas CogVLM2-19B has the lowest FPR. They both outperform WinCLIP, for which threshold-dependent metrics cannot be computed without a validation set. Overall, VLMs show high AUROC and AUPR, highlighting their potential as effective anomaly detectors. \textit{Enhanced-PatchCore} has a better F1 score than all VLMs except CogVLM-17B. There are large variations across VLMs, indicating the need for careful selection. CogVLM2-19B's higher AUROC and AUPR but worse F1 score suggest suboptimal thresholding, underscoring the challenge of effective threshold control in zero-shot VLMs. \textit{Enhanced-PatchCore}, even with limited nominal images, maintains competitiveness while offering the added advantage of pixel-level evaluation.

\begin{figure}[tb]
  \centering
\begin{subfigure}[b]{\textwidth}
    \centering
    \includegraphics[width=\textwidth,trim={0 0 0 10px},clip]{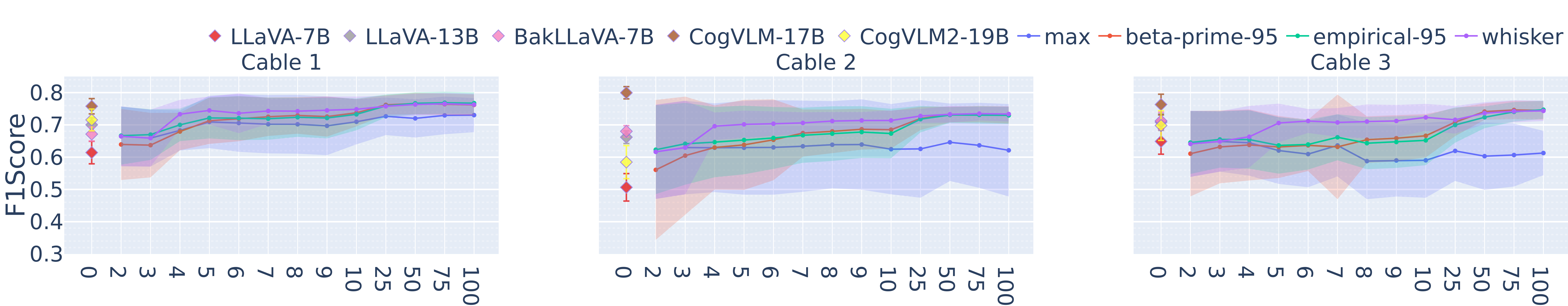}
    \caption{F1-Score as the number of train images increases on the x-axis.}
    \label{fig:f1score}
  \end{subfigure} 
  \begin{subfigure}[b]{\textwidth}
    \centering
    \includegraphics[width=\textwidth,trim={0 0 0 10px},clip]{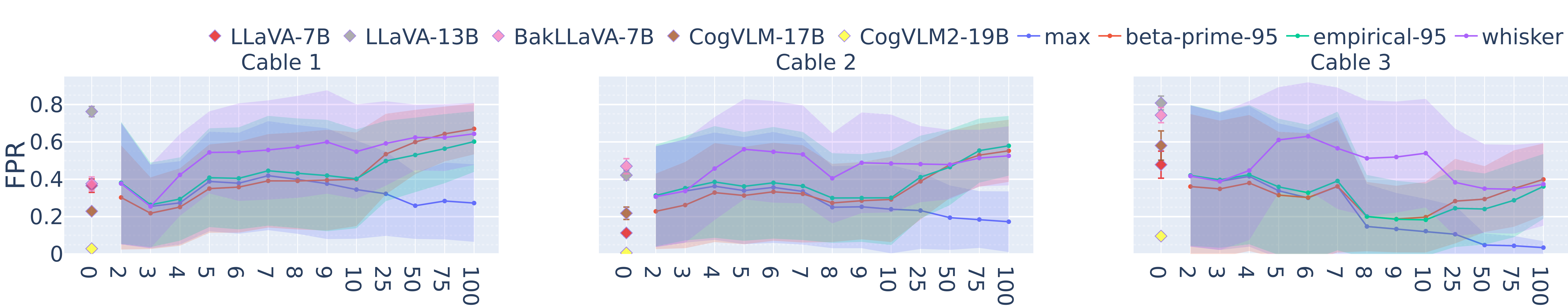}
    \caption{FPR as the number of train images increases on the x-axis.}
    \label{fig:fpr}
  \end{subfigure}
  \begin{subfigure}[b]{\textwidth}
    \centering
    \includegraphics[width=\textwidth,trim={0 0 0 30px},clip]{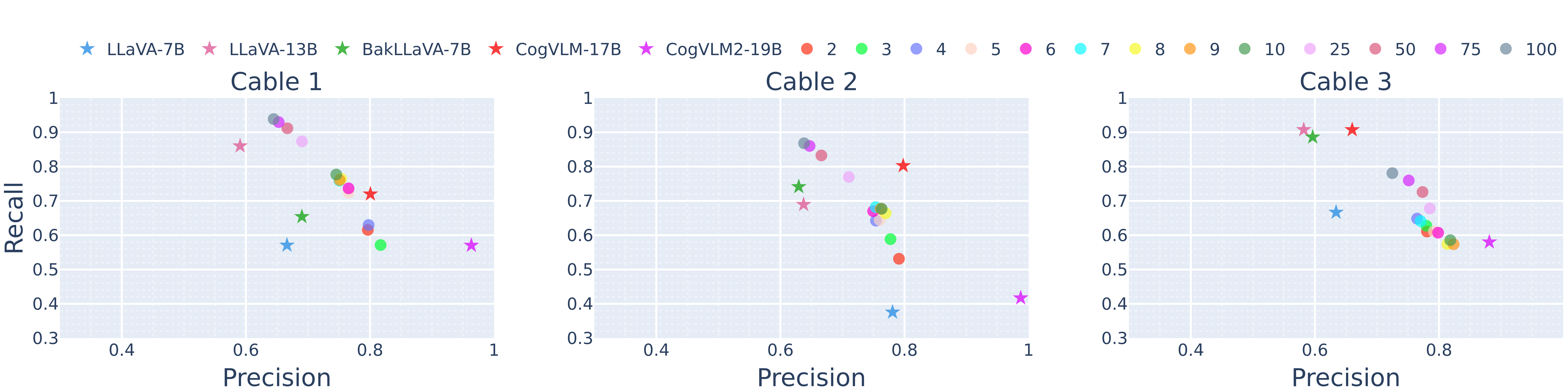}
    \caption{Precision vs Recall. (*) show VLMs in a zero-shot setting. (o) show \textit{Enhanced-PatchCore} with \textit{beta-prime-95} thresholding. Here, the colors represent the number of training images.}
    \label{fig:precvsrecall}
  \end{subfigure} 
  \caption{Image-level results of \textit{Enhanced-PatchCore} (few-/many-shot) with the thresholding strategies and conversational VLMs (zero-shot). (a) and (b) show the mean and standard deviation over all folds for F1-score and FPR for the three cables. The x-axis indicates the number of images in the training set. (c) shows mean precision vs mean recall over all folds.
  }\label{fig:original_results_patchcore_vlm}
\end{figure}

\paragraph{Performance variability in same category objects} \Cref{fig:original_results_patchcore_vlm} compares the threshold-dependent metrics on the \textit{CableInspect-AD} dataset for each of the three cables. While all models achieve relatively high mean F1-score values, their performance can significantly vary (\Cref{fig:f1score}) across folds and cables. These variations are particularly notable for cables 2 and 3, which, being older, contain artifacts like scratches and discoloration from natural wear. These artifacts were not considered as anomalies by the experts, posing a greater challenge. This underscores the uniqueness of our dataset, where objects of the same category can have a significantly variable appearance. Additionally, the performance varies across the folds because the test sets of each fold can differ in terms of anomaly types and grades (see \Cref{fig:anomalies_types_grades}). Consequently,
folds containing a higher proportion of harder-to-detect anomalies (e.g., long scratches) compared to easier ones might show lower performance. Furthermore, our analysis suggests that VLMs are more robust compared to other methods, showing more consistent performance across different folds and cables.

\paragraph{\textit{Enhanced-PatchCore} - thresholding without a validation set}
From \Cref{fig:f1score}, we observe that the model performs well despite thresholding on the training set. Specifically, the performances of \textit{Enhanced-PatchCore} in the few and many-shot settings employing various thresholding strategies show that the mean F1-score improves in most cases as the number of training images increases. Among the thresholding strategies, \textit{max}---which is the most sensitive to outliers in the memory bank---appears brittle, while \textit{whisker}, \textit{empirical-$95$} and \textit{beta-prime-$95$} seem to be more robust across the cables. Additionally, if we examine the precision-recall and FPR-FNR trade-offs, using the \textit{beta-prime-$95$} strategy as an example (\Cref{fig:fpr,fig:precvsrecall}), we observe that, overall, for cables 1 and 2, an increase in recall is accompanied by a decrease in precision, usually at the expense of an increase in FPR, accompanied by a decrease in FNR (i.e., 1 - Recall), as the number of training images increases. Moreover, increasing the number of images in the training set does not seem beneficial, as it increases the risk of including outliers in the memory bank. On the other hand, reducing the number of instances might result in a less diverse training set compared to the distribution of real-world nominal images.

\paragraph{Analysis of conversational VLMs} Table \ref{tab:performance_metrics} shows that the VLMs achieve promising results despite not using any training examples (zero-shot). Specifically, the CogVLM variants outperform the other baselines. In \Cref{fig:original_results_patchcore_vlm}, CogVLM-17B shows the highest mean F1-score with the lowest variance across folds, outperforming other baselines across all cables (\Cref{fig:f1score}), whereas CogVLM2-19B shows the lowest FPR. Despite these encouraging results, VLMs are challenged by many limitations. Notably, VLMs can exhibit limitations in instruction following \cite{huang2024language}, be prone to object hallucinations \cite{zhou2023analyzing}, generate factual errors about objects, attributes, and relations \cite{liu2024survey}, and be vulnerable to deceptive prompts \cite{qian2024easy}. Moreover, while conversational VLMs show promise in anomaly detection, their ability to accurately localize anomalies remains a challenge. To highlight some of these limitations, we present examples in Supplementary Material.

\begin{figure}[tb]
  \centering
\begin{subfigure}[b]{\textwidth}
    \centering
    \includegraphics[trim={0 0 0 30px},clip,width=\textwidth]{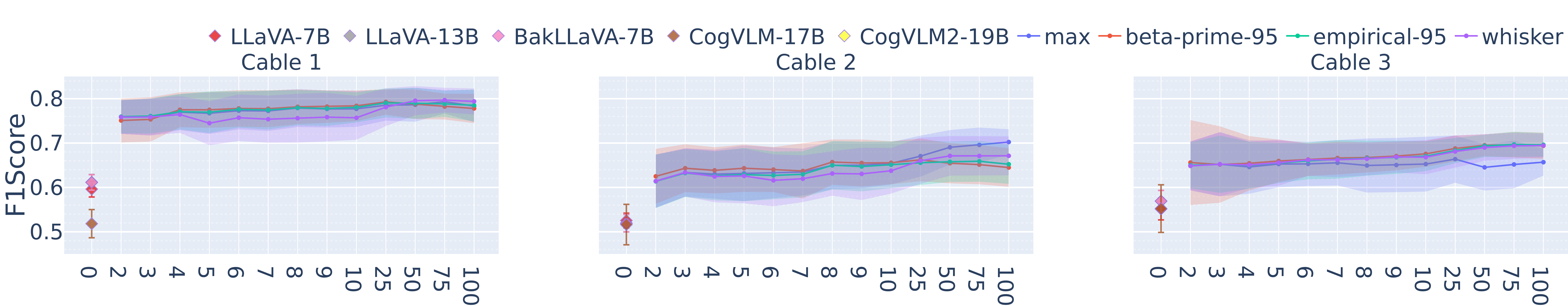}
    \caption{F1-Score as the number of train images increases on the x-axis.}
    \label{fig:f1score_tightcrop}
  \end{subfigure} 
  \begin{subfigure}[b]{\textwidth}
    \centering
    \includegraphics[width=\textwidth,trim={0 0 0 30px},clip,]{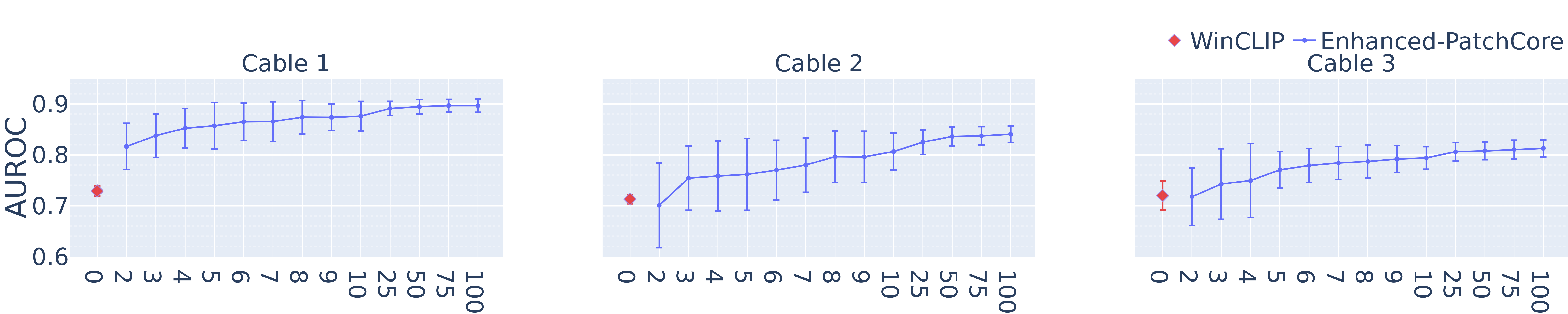}
    \caption{AUROC as the number of train images increases on the x-axis for \textit{Enhanced-PatchCore}.}
    \label{fig:auroc_tight_crop}
  \end{subfigure}
  \caption{Image-level results in zero-shot setting using conversational VLMs and WinCLIP, and, few-shot and many-shot using \textit{Enhanced-PatchCore} on \textit{CableInspect-AD\textunderscore cropped}. Mean and standard deviation over all folds are reported for the three cables. On the figures, the x-axis indicates the number of images in the training set. (a) shows F1-score. For \textit{Enhanced-PatchCore}, the metrics are computed using different thresholding strategies. (b) AUROC for \textit{Enhanced-PatchCore} and WinCLIP.}
  \label{fig:tight_crop_results_patchcore_vlm}
\end{figure}

\paragraph{Evaluating the impact of background removal} One possible reason for the high variability of the performances of \textit{Enhanced-PatchCore} is its sensitivity to variations in the background. Therefore, we evaluate the baseline models on a cropped version of \textit{CableInspect-AD}, namely \textit{CableInspect-AD\textunderscore cropped}, in which we retain only the central part of the cables. In \Cref{fig:tight_crop_results_patchcore_vlm}, \textit{Enhanced-PatchCore} shows lower variance in the F1-score across the different thresholding strategies while maintaining good performances on all cables. All thresholding strategies perform similarly, except for the \textit{max} strategy on cable 3. Thus, the extraction of the region of interest seems beneficial. Surprisingly, the performance of the conversational VLMs drop significantly. This could be attributed to the reduced view in the cropped version of the image, potentially making it more challenging for them.

In \Cref{fig:auroc_tight_crop}, we observe an increase in mean AUROC and a decrease in its variance as the number of training images increases, indicating that the choice of the training image in the few-shot setting can greatly influence the performance. However, the AUROC variance does not decrease when the background is retained. WinCLIP demonstrates enhancements in AUROC when excluding the background. Similar findings apply to AUPR. More details on metrics and visualizations are in Supplementary Material.

\begin{figure}[tb]
\centering
\includegraphics[trim={0 0 0 80px},clip,width=0.9\textwidth]{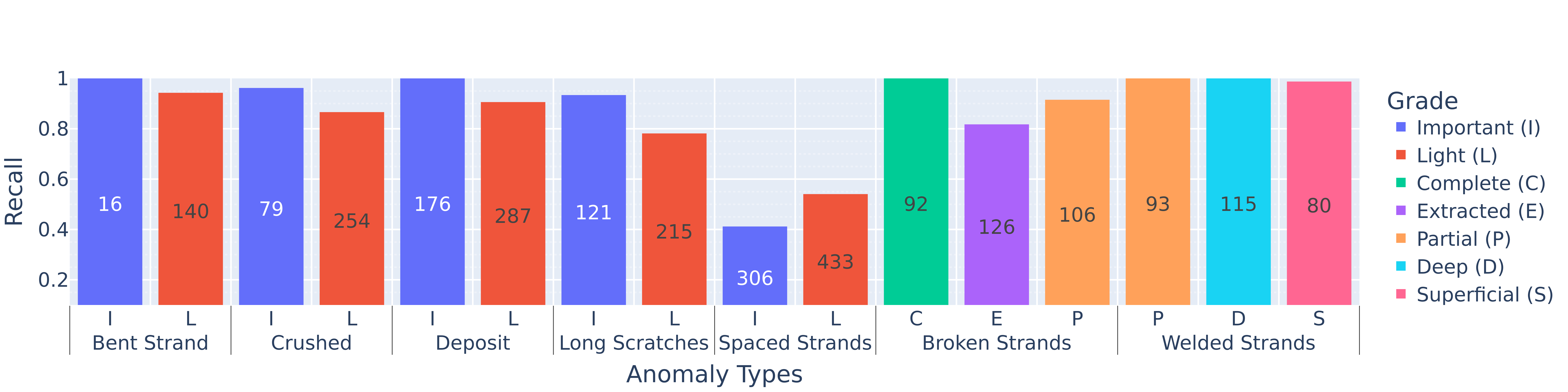}
\caption{CogVLM-17B image-level recall per anomaly types/grades (sample counts on bars).
}
\label{fig:anomaly_grade_results}
\end{figure}

\paragraph{Visual anomaly detection across different anomaly types and grades} Despite the promising performances demonstrated by the baseline models, all the models fail to detect all types/grades of anomalies. For instance, \Cref{fig:anomaly_grade_results} shows the recall of anomalies based on type and grade by CogVLM-17B on the whole \textit{CableInspect-AD} dataset. More pronounced anomaly types and grades such as \textit{bent strand (important)} and \textit{broken strand (complete)} are readily detected, whereas light and smaller anomalies such as \textit{spaced strands} and \textit{long scratches (light)} are prone to be overlooked. This highlights the importance of including multi-grade anomalies in the evaluation benchmark.

\paragraph{Anomaly Segmentation}
\textit{Enhanced-Patchcore} outperforms WinCLIP in the segmentation task on \textit{CableInspect-AD\textunderscore cropped}, with an AUPRO of $0.53$ $\pm$ $0.08$ compared to $0.27$ $\pm$ $0.06$ for WinCLIP. We apply thresholding strategies on anomaly maps generated by Enhanced-Patchcore to generate pixel-level predictions. We use a \textit{max} thresholding strategy for the segmentation results shown in \Cref{fig:segmentation_output} (more details are in the Supplementary material). The corresponding pixel-level metric, the Pixel-wise Overlap (PRO) score, averaged across all cables and folds, is 0.28 ± 0.09. \Cref{fig:segmentation_output} displays example outputs from \textit{Enhanced-Patchcore}, illustrating that the model effectively identifies larger anomalies but struggles with subtler ones. The rightmost image shows a nominal image where texture changes from wear are visible. These texture variations can distract the model adding complexity to the task. 

\begin{figure}[tb]
    \centering
    \begin{subfigure}[b]{\textwidth}
    \includegraphics[trim={0 0 0 0px},clip,width=\textwidth]
    {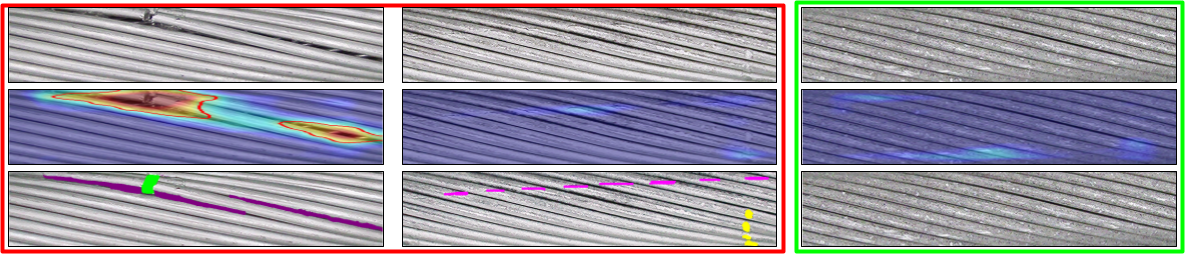}
    \end{subfigure}
    \caption{\textit{Enhanced-PatchCore} qualitative results for anomaly segmentation. The rightmost image is nominal (green); the rest show anomalies (red). The images (top row) and pixel-level prediction heatmaps with contours of detected anomalies using the \textit{max} thresholding strategy (middle row) are shown against ground truth masks (bottom row) from different cables. The bottom row shows the segmentation masks coloured based on the anomaly type. Some anomalies are easily detected (left column) whereas the others are difficult and are missed (middle column).}
    \label{fig:segmentation_output}
\end{figure}

\paragraph{Contribution} Our dataset demonstrates its unique strength through the comprehensive diversity of anomaly types and severity levels it captures. Specifically, it includes seven distinct types of anomalies, each with up to three levels of severity. This allows for a more in-depth evaluation within the targeted domain. Broader datasets, with lower anomaly diversity per category, may not fully capture the intricacies persistent in real-world applications. In addition, given the accelerating electrification of transportation, there is a growing need for reliable transmission facilities. Therefore, it is critical to develop VAD models that can specialize in such high-stakes applications. Our dataset meets this need by offering a focused evaluation framework that complements broader datasets.

\paragraph{Broad impact} The methodologies and insights derived from our focused study are adaptable to a wide range of anomaly detection scenarios. For instance, our experiments demonstrate that Vision-Language Models (VLMs) can be effectively utilized for zero-shot VAD tasks. However, we also find that no current model performs well across all anomaly types, particularly when detecting light-grade anomalies. This finding reveals the limitations of current models and provides a valuable direction for future research aimed at enhancing model performance in specialized applications.

\paragraph{Limitations} We acknowledge that this work has the following limitations. First, we aimed to create a dataset containing a comprehensive range of real-world anomalies. However, this resulted in a higher anomaly ratio than typically observed in real-world scenarios, where anomalies rarely occur. This can be addressed by analyzing the results with this variation in mind or, when necessary, by employing stratified sampling to adjust the anomaly ratio within the folds. Second, despite our efforts to provide a rich and diverse set of examples for effective model learning and evaluation, the dataset does not encompass every possible anomaly found on a cable in real-world settings, because the methodology for data creation may not fully capture all complexities encountered in real-world scenarios, such as the deposition of snow or bird droppings on the cable.

\paragraph{Ethical concerns} We do not anticipate significant risks of security threats or human rights violations in our work or its potential applications. However, while our work aims to improve system reliability, we remind researchers that deploying machine learning models for VAD in robotic power line inspection may miss anomalies, potentially compromising safety and public utility operations.

\section{Conclusion}
In this work, we introduce \textit{CableInspect-AD}, a novel anomaly detection dataset created and annotated by domain experts. We employ a k-fold evaluation to assess \textit{Enhanced-PatchCore} with multiple thresholding strategies, WinCLIP and open VLMs on the proposed dataset. We find that, in general, the baselines show promising results in detecting anomalies on the cables, but struggle to detect anomalies of certain types and grades. 
This presents an important challenge for the development of new models on this task and highlights the potential value of \textit{CableInspect-AD} as a resource for the broader AD community.
Furthermore, we highlight the potential of recent open VLMs in zero-shot anomaly detection, requiring minimal prompt engineering and no image preprocessing. Future work will aim to assess VLM's zero-shot capabilities to other anomaly tasks such as type/grade classification, localization, and segmentation.

\newpage

\section*{Acknowledgments and Disclosure of Funding}
This research was enabled in part by compute resources, software and technical help provided by Mila (mila.quebec). We thank Ali Harakeh and Pierre-Luc St-Charles from the Mila Applied Machine Learning Research Team (AMLRT) for fruitful discussions, brainstorming and feedback. We also thank Hydro-Québec and IREQ for their involvement throughout the project. The project received funding from Hydro-Québec and was further supported by governmental contributions from the Ministère de l'Économie, de l'Innovation et de l'Énergie (MEIE) and Innovation, Science and Economic Development Canada (ISED).

\bibliographystyle{abbrv}
\bibliography{references}

\begin{thebibliography}{10}

\bibitem{Recognizance-2}
Recognizance-2: Power lines detection.
\newblock \url{https://www.kaggle.com/competitions/recognizance-2}.
\newblock Accessed: 2023-11-28.

\bibitem{abdelfattah2020ttpla}
R.~Abdelfattah, X.~Wang, and S.~Wang.
\newblock Ttpla: An aerial-image dataset for detection and segmentation of transmission towers and power lines.
\newblock In {\em Proceedings of the Asian Conference on Computer Vision}, 2020.

\bibitem{agarwal2005geometric}
P.~K. Agarwal, S.~Har-Peled, K.~R. Varadarajan, et~al.
\newblock Geometric approximation via coresets.
\newblock {\em Combinatorial and computational geometry}, 52(1):1--30, 2005.

\bibitem{akcay2019ganomaly}
S.~Akcay, A.~Atapour-Abarghouei, and T.~P. Breckon.
\newblock Ganomaly: Semi-supervised anomaly detection via adversarial training.
\newblock In {\em Computer Vision--ACCV 2018: 14th Asian Conference on Computer Vision, Perth, Australia, December 2--6, 2018, Revised Selected Papers, Part III 14}, pages 622--637. Springer, 2019.

\bibitem{alhassan2020power}
A.~B. Alhassan, X.~Zhang, H.~Shen, and H.~Xu.
\newblock Power transmission line inspection robots: A review, trends and challenges for future research.
\newblock {\em International Journal of Electrical Power \& Energy Systems}, 118:105862, 2020.

\bibitem{bergmann2022beyond}
P.~Bergmann, K.~Batzner, M.~Fauser, D.~Sattlegger, and C.~Steger.
\newblock Beyond dents and scratches: Logical constraints in unsupervised anomaly detection and localization.
\newblock {\em International Journal of Computer Vision}, 130(4):947--969, 2022.

\bibitem{bergmann2019mvtec}
P.~Bergmann, M.~Fauser, D.~Sattlegger, and C.~Steger.
\newblock Mvtec ad--a comprehensive real-world dataset for unsupervised anomaly detection.
\newblock In {\em Proceedings of the IEEE/CVF conference on computer vision and pattern recognition}, pages 9592--9600, 2019.

\bibitem{bian2019monocular}
J.~Bian, X.~Hui, X.~Zhao, and M.~Tan.
\newblock A monocular vision--based perception approach for unmanned aerial vehicle close proximity transmission tower inspection.
\newblock {\em International Journal of Advanced Robotic Systems}, 16(1):1729881418820227, 2019.

\bibitem{cano2023vepl}
M.~Cano-Solis, J.~R. Ballesteros, and J.~W. Branch-Bedoya.
\newblock Vepl dataset: A vegetation encroachment in power line corridors dataset for semantic segmentation of drone aerial orthomosaics.
\newblock {\em Data}, 8(8):128, 2023.

\bibitem{cao2023segment}
Y.~Cao, X.~Xu, C.~Sun, Y.~Cheng, Z.~Du, L.~Gao, and W.~Shen.
\newblock Segment any anomaly without training via hybrid prompt regularization.
\newblock {\em arXiv preprint arXiv:2305.10724}, 2023.

\bibitem{cao2023towards}
Y.~Cao, X.~Xu, C.~Sun, X.~Huang, and W.~Shen.
\newblock Towards generic anomaly detection and understanding: Large-scale visual-linguistic model (gpt-4v) takes the lead.
\newblock {\em arXiv preprint arXiv:2311.02782}, 2023.

\bibitem{chen2023april}
X.~Chen, Y.~Han, and J.~Zhang.
\newblock April-gan: A zero-/few-shot anomaly classification and segmentation method for cvpr 2023 vand workshop challenge tracks 1\&2: 1st place on zero-shot ad and 4th place on few-shot ad.
\newblock {\em arXiv preprint arXiv:2305.17382}, 2023.

\bibitem{chen2023zero}
X.~Chen, Y.~Han, and J.~Zhang.
\newblock A zero-/few-shot anomaly classification and segmentation method for cvpr 2023 vand workshop challenge tracks 1\&2: 1st place on zero-shot ad and 4th place on few-shot ad.
\newblock {\em arXiv preprint arXiv:2305.17382}, 2023.

\bibitem{choi2019real}
H.~Choi, G.~Koo, B.~J. Kim, and S.~W. Kim.
\newblock Real-time power line detection network using visible light and infrared images.
\newblock In {\em 2019 International Conference on Image and Vision Computing New Zealand (IVCNZ)}, pages 1--6. IEEE, 2019.

\bibitem{de2022ptl}
F.~S. de~Oliveira, M.~de~Carvalho, P.~H.~T. Campos, A.~D.~S. Soares, A.~C. J{\'u}nior, and A.~C. R. D.~S. Quirino.
\newblock Ptl-ai furnas dataset: A public dataset for fault detection in power transmission lines using aerial images.
\newblock In {\em 2022 35th SIBGRAPI Conference on Graphics, Patterns and Images (SIBGRAPI)}, volume~1, pages 7--12. IEEE, 2022.

\bibitem{deng2023anovl}
H.~Deng, Z.~Zhang, J.~Bao, and X.~Li.
\newblock Anovl: Adapting vision-language models for unified zero-shot anomaly localization.
\newblock {\em arXiv preprint arXiv:2308.15939}, 2023.

\bibitem{gebru2021datasheets}
T.~Gebru, J.~Morgenstern, B.~Vecchione, J.~W. Vaughan, H.~Wallach, H.~D. Iii, and K.~Crawford.
\newblock Datasheets for datasets.
\newblock {\em Communications of the ACM}, 64(12):86--92, 2021.

\bibitem{gu2023anomalygpt}
Z.~Gu, B.~Zhu, G.~Zhu, Y.~Chen, M.~Tang, and J.~Wang.
\newblock Anomalygpt: Detecting industrial anomalies using large vision-language models.
\newblock {\em arXiv preprint arXiv:2308.15366}, 2023.

\bibitem{hamelin2019discrete}
P.~Hamelin, F.~Mirall{\`e}s, G.~Lambert, S.~Lavoie, N.~Pouliot, M.~Montfrond, and S.~Montambault.
\newblock Discrete-time control of linedrone: An assisted tracking and landing uav for live power line inspection and maintenance.
\newblock In {\em 2019 International Conference on Unmanned Aircraft Systems (ICUAS)}, pages 292--298. IEEE, 2019.

\bibitem{huang2024language}
S.~Huang, L.~Dong, W.~Wang, Y.~Hao, S.~Singhal, S.~Ma, T.~Lv, L.~Cui, O.~K. Mohammed, B.~Patra, et~al.
\newblock Language is not all you need: Aligning perception with language models.
\newblock {\em Advances in Neural Information Processing Systems}, 36, 2024.

\bibitem{jeong2023winclip}
J.~Jeong, Y.~Zou, T.~Kim, D.~Zhang, A.~Ravichandran, and O.~Dabeer.
\newblock Winclip: Zero-/few-shot anomaly classification and segmentation.
\newblock In {\em Proceedings of the IEEE/CVF Conference on Computer Vision and Pattern Recognition}, pages 19606--19616, 2023.

\bibitem{kirillov2023segment}
A.~Kirillov, E.~Mintun, N.~Ravi, H.~Mao, C.~Rolland, L.~Gustafson, T.~Xiao, S.~Whitehead, A.~C. Berg, W.-Y. Lo, et~al.
\newblock Segment anything.
\newblock {\em arXiv preprint arXiv:2304.02643}, 2023.

\bibitem{lee2017weakly}
S.~J. Lee, J.~P. Yun, H.~Choi, W.~Kwon, G.~Koo, and S.~W. Kim.
\newblock Weakly supervised learning with convolutional neural networks for power line localization.
\newblock In {\em 2017 IEEE Symposium Series on Computational Intelligence (SSCI)}, pages 1--8. IEEE, 2017.

\bibitem{lee2022anovit}
Y.~Lee and P.~Kang.
\newblock Anovit: Unsupervised anomaly detection and localization with vision transformer-based encoder-decoder.
\newblock {\em IEEE Access}, 10:46717--46724, 2022.

\bibitem{li2021cutpaste}
C.-L. Li, K.~Sohn, J.~Yoon, and T.~Pfister.
\newblock Cutpaste: Self-supervised learning for anomaly detection and localization.
\newblock In {\em Proceedings of the IEEE/CVF conference on computer vision and pattern recognition}, pages 9664--9674, 2021.

\bibitem{li2024musc}
X.~Li, Z.~Huang, F.~Xue, and Y.~Zhou.
\newblock Musc: Zero-shot industrial anomaly classification and segmentation with mutual scoring of the unlabeled images.
\newblock {\em arXiv preprint arXiv:2401.16753}, 2024.

\bibitem{lin2024evaluating}
Z.~Lin, D.~Pathak, B.~Li, J.~Li, X.~Xia, G.~Neubig, P.~Zhang, and D.~Ramanan.
\newblock Evaluating text-to-visual generation with image-to-text generation.
\newblock 2024.

\bibitem{liu2023improved}
H.~Liu, C.~Li, Y.~Li, and Y.~J. Lee.
\newblock Improved baselines with visual instruction tuning.
\newblock {\em arXiv preprint arXiv:2310.03744}, 2023.

\bibitem{liu2024survey}
H.~Liu, W.~Xue, Y.~Chen, D.~Chen, X.~Zhao, K.~Wang, L.~Hou, R.~Li, and W.~Peng.
\newblock A survey on hallucination in large vision-language models.
\newblock 2024.

\bibitem{liu2023deep}
J.~Liu, G.~Xie, J.~Wang, S.~Li, C.~Wang, F.~Zheng, and Y.~Jin.
\newblock Deep industrial image anomaly detection: A survey.
\newblock {\em arXiv e-prints}, pages arXiv--2301, 2023.

\bibitem{madaan2017wire}
R.~Madaan, D.~Maturana, and S.~Scherer.
\newblock Wire detection using synthetic data and dilated convolutional networks for unmanned aerial vehicles.
\newblock In {\em 2017 IEEE/RSJ International Conference on Intelligent Robots and Systems (IROS)}, pages 3487--3494. IEEE, 2017.

\bibitem{openai2023gpt4v}
OpenAI.
\newblock Gpt-4v(ision) system card.
\newblock OpenAI, September 25 2023.
\newblock Available at: \url{https://openai.com/index/gpt-4v-system-card/} (Accessed: 30 November 2023).

\bibitem{prates2019insulator}
R.~M. Prates, R.~Cruz, A.~P. Marotta, R.~P. Ramos, E.~F. Simas~Filho, and J.~S. Cardoso.
\newblock Insulator visual non-conformity detection in overhead power distribution lines using deep learning.
\newblock {\em Computers \& Electrical Engineering}, 78:343--355, 2019.

\bibitem{qian2024easy}
Y.~Qian, H.~Zhang, Y.~Yang, and Z.~Gan.
\newblock How easy is it to fool your multimodal llms? an empirical analysis on deceptive prompts.
\newblock 2024.

\bibitem{richard2019lineranger}
P.-L. Richard, N.~Pouliot, F.~Morin, M.~Lepage, P.~Hamelin, M.~Lagac, A.~Sartor, G.~Lambert, and S.~Montambault.
\newblock Lineranger: Analysis and field testing of an innovative robot for efficient assessment of bundled high-voltage powerlines.
\newblock In {\em 2019 International Conference on Robotics and Automation (ICRA)}, pages 9130--9136. IEEE, 2019.

\bibitem{roth2022towards}
K.~Roth, L.~Pemula, J.~Zepeda, B.~Sch{\"o}lkopf, T.~Brox, and P.~Gehler.
\newblock Towards total recall in industrial anomaly detection.
\newblock In {\em Proceedings of the IEEE/CVF Conference on Computer Vision and Pattern Recognition}, pages 14318--14328, 2022.

\bibitem{rudolph2022fully}
M.~Rudolph, T.~Wehrbein, B.~Rosenhahn, and B.~Wandt.
\newblock Fully convolutional cross-scale-flows for image-based defect detection.
\newblock In {\em Proceedings of the IEEE/CVF Winter Conference on Applications of Computer Vision}, pages 1088--1097, 2022.

\bibitem{santos2023optimizing}
J.~Santos, T.~Tran, and O.~Rippel.
\newblock Optimizing patchcore for few/many-shot anomaly detection.
\newblock {\em arXiv preprint arXiv:2307.10792}, 2023.

\bibitem{shi2016uniad}
X.~Shi, B.~Cui, G.~Dobbie, and B.~C. Ooi.
\newblock Uniad: A unified ad hoc data processing system.
\newblock {\em ACM Transactions on Database Systems (TODS)}, 42(1):1--42, 2016.

\bibitem{song2021anoseg}
J.~Song, K.~Kong, Y.-I. Park, S.-G. Kim, and S.-J. Kang.
\newblock Anoseg: anomaly segmentation network using self-supervised learning.
\newblock {\em arXiv preprint arXiv:2110.03396}, 2021.

\bibitem{tao2018detection}
X.~Tao, D.~Zhang, Z.~Wang, X.~Liu, H.~Zhang, and D.~Xu.
\newblock Detection of power line insulator defects using aerial images analyzed with convolutional neural networks.
\newblock {\em IEEE transactions on systems, man, and cybernetics: systems}, 50(4):1486--1498, 2018.

\bibitem{tomaszewski2018collection}
M.~Tomaszewski, B.~Ruszczak, and P.~Michalski.
\newblock The collection of images of an insulator taken outdoors in varying lighting conditions with additional laser spots.
\newblock {\em Data in brief}, 18:765--768, 2018.

\bibitem{tsai2022multi}
C.-C. Tsai, T.-H. Wu, and S.-H. Lai.
\newblock Multi-scale patch-based representation learning for image anomaly detection and segmentation.
\newblock In {\em Proceedings of the IEEE/CVF Winter Conference on Applications of Computer Vision}, pages 3992--4000, 2022.

\bibitem{vieira2021stn}
A.~L.~B. Vieira-e Silva, H.~de~Castro~Felix, T.~de~Menezes~Chaves, F.~P.~M. Simoes, V.~Teichrieb, M.~M. dos Santos, H.~da~Cunha~Santiago, V.~A.~C. Sgotti, and H.~B. D. T.~L. Neto.
\newblock Stn plad: A dataset for multi-size power line assets detection in high-resolution uav images.
\newblock In {\em 2021 34th SIBGRAPI Conference on Graphics, Patterns and Images (SIBGRAPI)}, pages 215--222. IEEE, 2021.

\bibitem{vieira2023insplad}
A.~L.~B. Vieira~e Silva, H.~de~Castro~Felix, F.~P.~M. Sim{\~o}es, V.~Teichrieb, M.~dos Santos, H.~Santiago, V.~Sgotti, and H.~Lott~Neto.
\newblock Insplad: A dataset and benchmark for power line asset inspection in uav images.
\newblock {\em International journal of remote sensing}, 44(23):7294--7320, 2023.

\bibitem{wang2021student}
G.~Wang, S.~Han, E.~Ding, and D.~Huang.
\newblock Student-teacher feature pyramid matching for anomaly detection.
\newblock {\em arXiv preprint arXiv:2103.04257}, 2021.

\bibitem{wang2023cogvlm}
W.~Wang, Q.~Lv, W.~Yu, W.~Hong, J.~Qi, Y.~Wang, J.~Ji, Z.~Yang, L.~Zhao, X.~Song, J.~Xu, B.~Xu, J.~Li, Y.~Dong, M.~Ding, and J.~Tang.
\newblock Cogvlm: Visual expert for pretrained language models.
\newblock 2023.

\bibitem{wyatt2022anoddpm}
J.~Wyatt, A.~Leach, S.~M. Schmon, and C.~G. Willcocks.
\newblock Anoddpm: Anomaly detection with denoising diffusion probabilistic models using simplex noise.
\newblock In {\em Proceedings of the IEEE/CVF Conference on Computer Vision and Pattern Recognition}, pages 650--656, 2022.

\bibitem{yang2020dfr}
J.~Yang, Y.~Shi, and Z.~Qi.
\newblock Dfr: Deep feature reconstruction for unsupervised anomaly segmentation.
\newblock {\em arXiv preprint arXiv:2012.07122}, 2020.

\bibitem{yetgin2017ground}
{\"O}.~E. Yetgin, {\"O}.~N. Gerek, and {\"O}.~Nezih.
\newblock Ground truth of powerline dataset (infrared-ir and visible light-vl).
\newblock {\em Mendeley Data}, 8(9), 2017.

\bibitem{yoon2021self}
J.~Yoon, K.~Sohn, C.-L. Li, S.~O. Arik, C.-Y. Lee, and T.~Pfister.
\newblock Self-supervise, refine, repeat: Improving unsupervised anomaly detection.
\newblock {\em arXiv preprint arXiv:2106.06115}, 2021.

\bibitem{you2022adtr}
Z.~You, K.~Yang, W.~Luo, L.~Cui, Y.~Zheng, and X.~Le.
\newblock Adtr: Anomaly detection transformer with feature reconstruction.
\newblock In {\em International Conference on Neural Information Processing}, pages 298--310. Springer, 2022.

\bibitem{zavrtanik2021draem}
V.~Zavrtanik, M.~Kristan, and D.~Sko{\v{c}}aj.
\newblock Draem-a discriminatively trained reconstruction embedding for surface anomaly detection.
\newblock In {\em Proceedings of the IEEE/CVF International Conference on Computer Vision}, pages 8330--8339, 2021.

\bibitem{zhang2023exploring}
J.~Zhang, X.~Chen, Z.~Xue, Y.~Wang, C.~Wang, and Y.~Liu.
\newblock Exploring grounding potential of vqa-oriented gpt-4v for zero-shot anomaly detection.
\newblock {\em arXiv preprint arXiv:2311.02612}, 2023.

\bibitem{zhou2023analyzing}
Y.~Zhou, C.~Cui, J.~Yoon, L.~Zhang, Z.~Deng, C.~Finn, M.~Bansal, and H.~Yao.
\newblock Analyzing and mitigating object hallucination in large vision-language models.
\newblock 2023.

\bibitem{zou2022spot}
Y.~Zou, J.~Jeong, L.~Pemula, D.~Zhang, and O.~Dabeer.
\newblock Spot-the-difference self-supervised pre-training for anomaly detection and segmentation.
\newblock In {\em European Conference on Computer Vision}, pages 392--408. Springer, 2022.

\end{thebibliography}
\newpage

\appendix

\setcounter{figure}{0}
\setcounter{table}{0}

\renewcommand{\thesubsection}{\Alph{subsection}}
\renewcommand\thefigure{S\arabic{figure}}
\renewcommand\thetable{S\arabic{table}}

\section*{Supplementary Material for \\ \textit{CableInspect-AD: An Expert-Annotated Anomaly Detection Dataset}}

\begin{figure}[ht!]
    \centering
    \includegraphics[width=\linewidth]{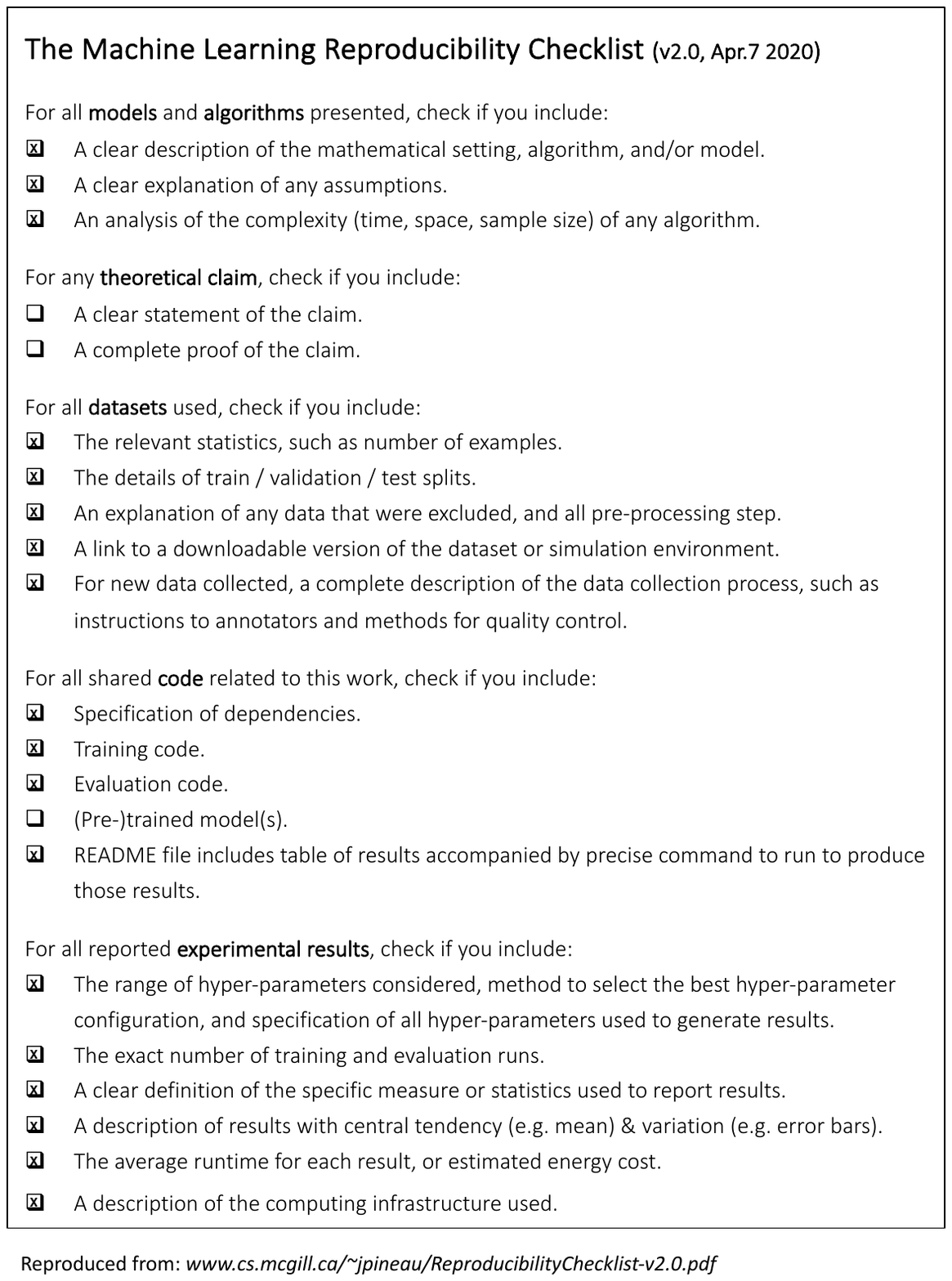}
    \caption{Reproducibility checklist}
    \label{fig:ml_checklist}
\end{figure}

We provide links to the dataset and the code repository for reproducibility in subsection \ref{sec:links} along with the author statement \ref{sec:author_statement}. The detailed dataset documentation and intended uses in the form of a datasheet for datasets \cite{gebru2021datasheets} are available in subsection \ref{sec:appendix-datasheet}. We also include an ML reproducibility checklist (see \ref{fig:ml_checklist}). 

In the following subsections, we present the dataset creation and annotation process (see \ref{sec:dataset_creation_annotation}), the dataset partitioning using k-fold cross-validation (see \ref{sec:dataset_partitioning}), a description of the thresholding strategies used (see \ref{sec:thresholding_strategies}), more details on the background removal procedure (see \ref{sec:background_removal}), implementation details (see \ref{sec:implementation_details}), threshold-independent metrics on \textit{Enhanced-PatchCore} (see \ref{sec:threshold_independent_metrics_patchcore}), and a few qualitative examples obtained with VLMs (see \ref{sec:qualitative_examples_vlms}).

\subsection{Dataset and code access links}
\label{sec:links}
The project website link associated with the paper is the following: \url{https://mila-iqia.github.io/cableinspect-ad/}.
\begin{itemize}
    \item \textbf{Dataset}: The dataset can be accessed via the \textit{Data} icon/hyperlink in the project website: \url{https://mila-iqia.github.io/cableinspect-ad/}. The dataset is hosted and maintained by the authors. For more information, please refer to the \textit{Distribution} and \textit{Maintenance} subsections of the datasheet provided in \ref{sec:appendix-datasheet}. The annotations are in the COCO format. We provide detailed explanations on how the dataset can be read in the code repository.
    \item \textbf{Code}: The link to the code repository is the following: \url{https://github.com/mila-iqia/cableinspect-ad-code}. The repository includes the code necessary to process the dataset, as well as the code required to reproduce all the experiments presented in the paper.
\end{itemize}

\subsection{Author statement}
\label{sec:author_statement}
We, the authors of the submitted paper titled \textit{CableInspect-AD: An Expert-Annotated Anomaly Detection Dataset}, hereby affirm the following:
 \begin{itemize}
\item \textbf{Responsibility for Content}: We bear full responsibility for the content of this paper, including any potential violation of rights or legal issues arising from the use or distribution of the dataset described in our submission. 
\item  \textbf{Data License Confirmation}: The dataset developed is licensed under Attribution NonCommercial ShareAlike 4.0 International License (CC BY-NC-SA 4.0).
 \end{itemize}

\subsection{Dataset creation and annotation}
\label{sec:dataset_creation_annotation}

\begin{figure}[ht!]
  \centering
  \begin{subfigure}[b]{0.3\textwidth}
    \centering
    \includegraphics[height=0.56\textwidth]{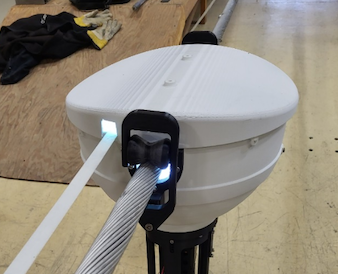}
    \caption{}
    \label{fig:image_a}
  \end{subfigure}
  \begin{subfigure}[b]{0.3\textwidth}
    \centering
    \includegraphics[width=\textwidth]{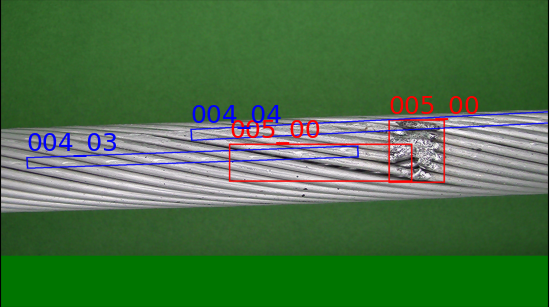}
    \caption{}
    \label{fig:image_b}
  \end{subfigure}
  \begin{subfigure}[b]{0.3\textwidth}
    \centering
    \includegraphics[width=\textwidth]{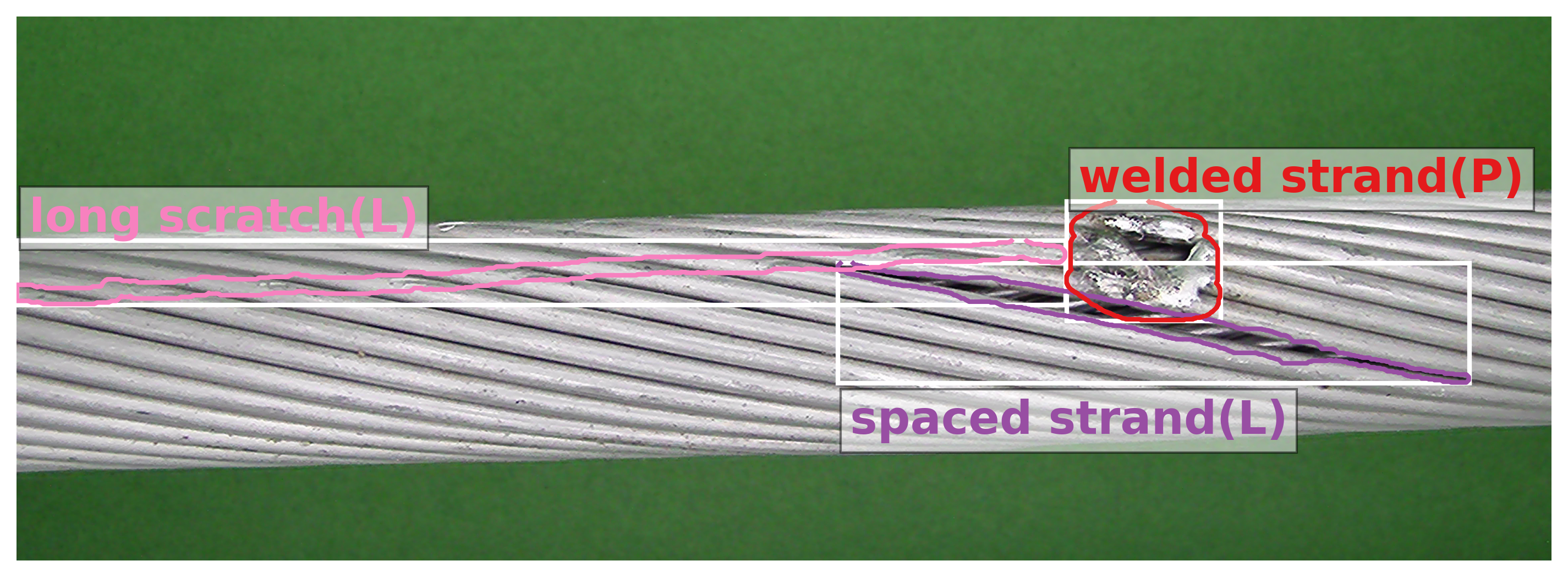}
    \caption{}
    \label{fig:image_c}
  \end{subfigure}
  \caption{Image acquisition process.
  Image (a) shows the prototype of the apparatus used to control the background and the lighting during the acquisition phase. 
  Image (b) shows an example of \textit{CableInspect-AD} after post-processing and annotation. As we can see, a green band is added at the bottom of the image to cover the tape used for marking the location of the different anomalies, which was used during the annotation process. This measure aims to prevent the model from exploiting this information. The image has defects with more than one anomaly type. The defect labeled as 005\textunderscore 00 has multiple anomaly types: the left side of the defect is a light-spaced strand, while the right side is a partially welded strand. Furthermore, within this image, two additional defects can be identified: 004\textunderscore 03 and 004\textunderscore 04, both of which are light long scratches. Image (c) shows an example of pixel-level annotation.}
  \label{fig:acquisition_uc1}
\end{figure}

\Cref{tab:anomalies_descriptions} presents the anomalies annotation guidelines and \Cref{fig:acquisition_uc1} presents the image acquisition process. The dataset underwent five iterative rounds of review and feedback, allowing the experts to reach a consensus. This process ensured that the final version was both reliable and reflective of real-world conditions. While very light anomalies, such as light deposits and scratches, might have been missed, the experts agreed these are not critical, as they would not require immediate repair in a real-world scenario and might even go undetected by experts. All mild and severe cases were thoroughly annotated. We did not quantify the annotation process’ performance, as it was conducted in a consensus-driven, iterative manner until an agreement was reached. 

\begin{table}[ht!]
\caption{Anomaly types and grades annotation guidelines.}
\resizebox{\textwidth}{!}{\small
\begin{tabular}{@{}llp{80mm}@{}}
\toprule
Anomaly Type                    & Grade & Description \\ \midrule
\multirow{3}{*}{Welded strand} & Superficial   & Each strand is identifiable. \\ \cmidrule(l){2-3} 
                                & Partial       & Strands are fused together. \\ \cmidrule(l){2-3} 
                                & Deep          & A strand is completely disconnected by the fusion. \\ \midrule
\multirow{3}{*}{Broken strand} & Partial       & The strand is modified but still connected. \\ \cmidrule(l){2-3} 
                                & Complete      & The strand is completely cut but still in place. \\ \cmidrule(l){2-3}
                                & Extracted     & The strand is cut; part of it is seen outside of the cable. \\ \midrule
\multirow{2}{*}{Spaced strand} & Light         & Slightly spaced, the next layer of strands cannot be seen. \\ \cmidrule(l){2-3} 
                                & Important     & Next layer of strands can be seen. \\ \midrule
\multirow{2}{*}{Bent strand}    & Light         & Distortion smaller than the width of a strand. \\ \cmidrule(l){2-3} 
                                & Important     & Distortion bigger than the width of a strand. \\ \midrule
\multirow{2}{*}{Crushed}        & Light         & Crushed part is smaller than the width of a strand. \\ \cmidrule(l){2-3} 
                                & Important     & Crushed part is bigger than the width of a strand. \\ \midrule
\multirow{2}{*}{Long scratch} & Light         & Scratch width smaller than 1/3 of a strand width. \\ \cmidrule(l){2-3} 
                                & Important     & Scratch width bigger than 1/3 of a strand width. \\ \midrule
\multirow{2}{*}{Deposit}        & Light         & Deposit is smaller than the width of a strand. \\ \cmidrule(l){2-3} 
                                & Important     & Deposit is bigger than the width of a strand. \\ \bottomrule
\label{tab:anomalies_descriptions}
\end{tabular}
}
\end{table}

\subsection{Dataset partitioning using k-fold cross-validation}
\label{sec:dataset_partitioning}

\begin{figure}[tb]
\centering
\includegraphics[width=0.80\textwidth,trim={0 10px 0 5px},clip]{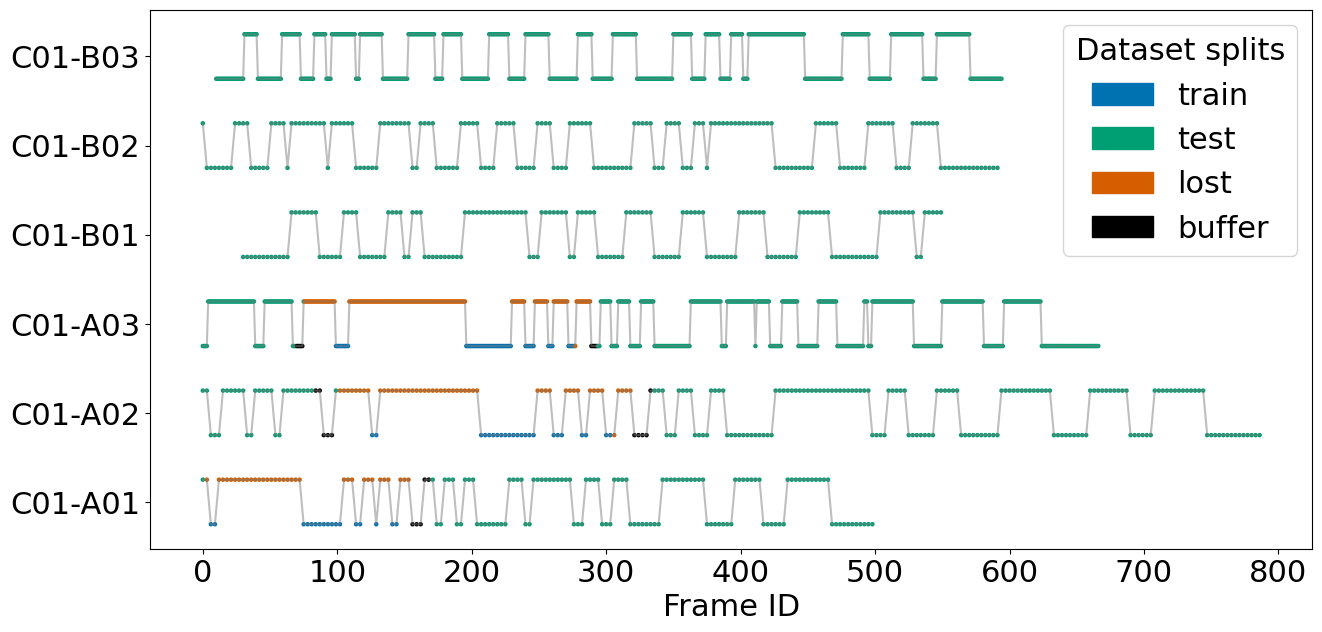}
\caption{Example of one split in the k-fold for cable 1 (C01). Cable 1 has two sides, A and B, and three passes are done to capture the images 01, 02, and 03. Variations in the number of frames across cables result from slight fluctuations in the apparatus speed during manual acquisition and the fact that the cables are not of the same length. Additionally, initial frames showing poor quality were excluded from the dataset. Here, the lines represent the cable videos, and each dot within the lines represents a frame. The nominal images are at the lower level, while anomalous images are at the upper level (peaks) of the lines. Only nominal images are in the training set. Images in the training section that are not part of the training set are labeled as \emph{lost}. Additionally, images associated with the two buffers are excluded. All remaining images, including those on the opposite side of the cable, constitute the test set. These images include both nominal and anomalous images.}
\label{fig:kfold_split}
\end{figure}

The power line cable dataset is split into train and test sets using a k-fold sampling strategy based on defect identifiers. We consider each cable side independently, as anomalies with the same identifier do not often occur on both sides of the cable. Moreover, when an anomaly appears on both sides, its visual characteristics differ depending on the point of view.

To generate a fold, we start by randomly selecting a defect identifier and retrieving its corresponding images on the same cable side. This marks the beginning of the training section. In total, 100 nominal images following this defect are included in the training set. We included 100 images to have a small training set in the same order of magnitude as the popular MVTec AD benchmark. The next defect identifier (following these 100 nominal images) marks the end of the training section. Images between the 100\textsuperscript{th} image and the next defect identifier are discarded. To remove any overlap between cable sections in training and test sets, we use buffers before and after the training section (see \Cref{fig:kfold_split}).

This process is repeated $k$ times, sequentially selecting defect identifiers from an ordered list spanning the entire cable length for each cable. Although each fold contains a constant number of training images (100), the number of test images and the anomaly ratios vary across folds.

\subsection{Thresholding strategies}
\label{sec:thresholding_strategies}

\begin{figure}[!t]
\centering
\includegraphics[width=0.9\textwidth,trim={0 5px 0 10px},clip]{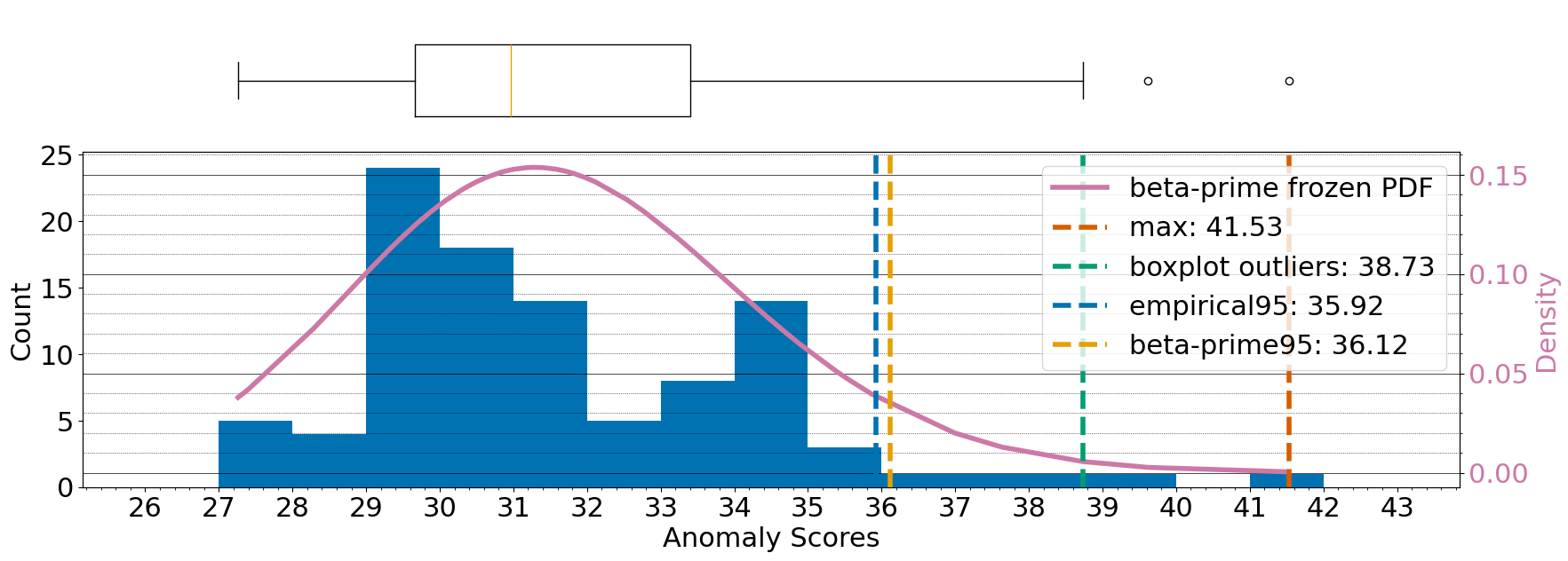}
\caption{Thresholding strategies. The histogram shows an example of the empirical distribution of anomaly scores within the training set in a fold, with markers indicating the thresholds of the four thresholding strategies. Positioned above the histogram is the boxplot, which is used to set the \emph{whisker} threshold.  Additionally, superimposed on the histogram is the beta-prime fit of the training anomaly score distribution, used to set the \emph{beta-prime} threshold. $\alpha = 95$ percentile is used to set both the \emph{beta-prime} and \emph{empirical} thresholds. Here, PDF stands for Probability Density Function.}
\label{fig:all_thresholds_histogram}
\end{figure}

To generate a threshold for the threshold-dependent metrics, we experiment with four thresholding strategies (see \Cref{fig:all_thresholds_histogram}):
\begin{itemize}
\item \textbf{\textit{Max}}: The maximum anomaly score of the empirical distribution obtained from the training data is chosen as the threshold. Given the assumption that the training data contains only nominal images, this threshold should be lower than the scores associated with anomalies in the test set. However, in practice, the \textit{max} strategy is sensitive to outliers in the training data. 
\item \textbf{Outliers from a boxplot (\textit{whisker})}: In a box-and-whisker plot, the points beyond the whiskers are considered outliers. To detect anomalies, the point at the upper quartile whisker is selected as the threshold. This corresponds to the largest anomaly score that is within $1.5 \times IQR$ above the third quartile ($Q3$), where $IQR$ is the interquartile range $Q3 - Q1$.
\item \textbf{Percentile estimation from empirical distribution (\textit{empirical-$\alpha$)}}: The observed anomaly scores are sorted, and the value corresponding to the $\alpha$ percentile is chosen as the threshold.
\item \textbf{Percentile estimation from parametric distribution (\textit{beta-prime-$\alpha$})}: A beta-prime distribution is fit to the anomaly scores, and the value at $\alpha$ percentile is chosen as the threshold. By using a prior on the distribution family of the score, we expect the algorithm to be more robust in the low-data regime.
\end{itemize}

\subsection{Background removal}
\label{sec:background_removal}

\begin{figure}[htb!]
  \centering
  \begin{subfigure}[b]{0.3\textwidth}
    \centering
    \includegraphics[width=\textwidth]{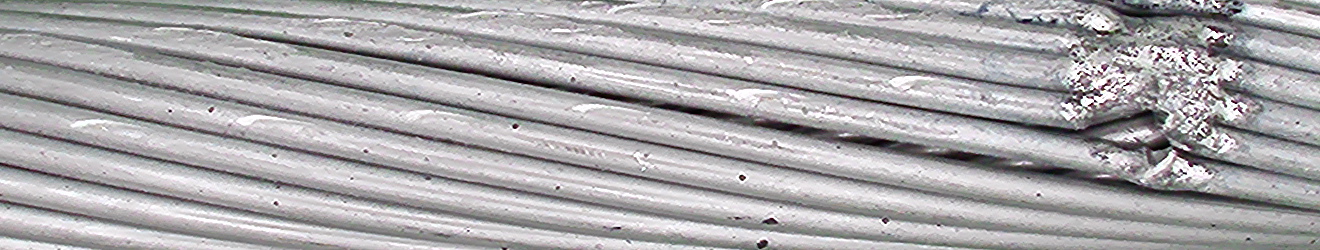}
    \caption{}
  \end{subfigure}
  \caption{The image shows an example of \textit{CableInspect-AD\textunderscore cropped}.}
  \label{fig:tight_crop_image}
\end{figure}

We create \textit{CableInspect-AD\textunderscore cropped} dataset, containing the images with the background removed, keeping only the central part of the cables. The dataset was generated by extracting a central band of size $224 \times 1120$ as shown in \cref{fig:tight_crop_image}. During the ROI extraction, 696 anomalous images (typically the ones where anomalies extend outside the cable) out of the original 2639 become nominal, resulting in a dataset containing 4798 images: 2855 nominal and 1943 anomalous. Specifically, we lose ten unique anomalies, corresponding to six \textit{broken strands (extracted)}, two \textit{bent strands (light)}, one \textit{broken strand (complete)}, and one \textit{spaced strand (light)}. Furthermore, all the remaining 183 anomalies lose some of their views.

\subsection{Implementation details}
\label{sec:implementation_details}

\begin{table}[ht!]
\centering
\caption{Main characteristics of the Vision-Language Models (VLMs) used in this work. The table provides details on each model, including the vision encoder (with its corresponding input image resolution in pixels), the Large Language Model (LLM) backbone, the multimodal alignment strategy, and the name of the weights used for inference from the \textit{transformers} library (i.e., \textit{HuggingFace} platform). For CogVLM variants, the \textit{Visual Expert Module} refers to the vision-specific layers incorporated inside the LLM architecture to enhance multimodal alignment via deep fusion.}
\label{tab:vlms}
\resizebox{\textwidth}{!}{
\begin{tabular}{@{}p{2.5cm}p{2.5cm}p{2.5cm}p{3.5cm}p{3.7cm}@{}}
\toprule
Model & Vision Encoder & LLM Backbone & Multimodal Alignment & \textit{HuggingFace} Weights \\ 
\midrule
LLaVA-1.5-7B & CLIP-ViT-L/14 ($336^2$) & Vicuna-1.5-7B & MLP Projector & \textit{llava-hf/llava-1.5-7b-hf} \\
\addlinespace
LLaVA-1.5-13B & CLIP-ViT-L/14 ($336^2$) & Vicuna-1.5-13B & MLP Projector & \textit{llava-hf/llava-1.5-13b-hf} \\
\addlinespace
BakLLaVA-7B & CLIP-ViT-L/14 ($336^2$) & Mistral-7B & MLP Projector & \textit{llava-hf/bakLlava-v1-hf} \\
\addlinespace
CogVLM-17B & EVA02-CLIP-E ($490^2$) & Vicuna-1.5-7B & MLP Projector and a \hspace{1cm} Visual Expert Module & \textit{THUDM/cogvlm-chat-hf} \\ 
\addlinespace
CogVLM2-19B & EVA02-CLIP-E ($1344^2$) & LLaMA-3-8B-Instruct & MLP Projector and a \hspace{1cm} Visual Expert Module & \textit{THUDM/cogvlm2-llama3-chat-19B} \\ 
\bottomrule
\end{tabular}
}
\end{table}

\textit{Enhanced-PatchCore} was developed on top of PatchCore from \textit{anomalib}\footnote{\url{https://github.com/openvinotoolkit/anomalib}} implementation with default hyperparameters. We sampled $n = {2, 3, 4, 5, 6, 7, 8, 9, 10}$ images for few-shot and $n={25, 50, 75, 100}$ images for many-shot experiments as part of the training set, excluding the zero-shot scenario as it requires at least two images to constitute a memory bank in our enhanced version. We applied individual models for each cable to account for their distinct characteristics.

For VLMs we used the implementations from the \textit{transformers}\footnote{\url{https://github.com/huggingface/transformers}} library. LLaVA 1.5 is a recent iteration of the original LLaVA, with improvements in multimodal alignment by replacing the original linear projector with a two-layer MLP projector, as well as integrating academic task-oriented data into its training pipeline. BakLLaVA uses the same architecture as LLaVA 1.5, but replaces the Vicuna Large Language Model (LLM) backbone with a Mistral backbone instead. CogVLM models integrate a visual expert module inside the LLM backbone, enabling deeper fusion between the vision and language embeddings. Furthermore, they take higher input resolutions in their vision encoders and leverage a much larger pre-trained vision encoder. \Cref{tab:vlms} highlights these differences. 

For all VLMs model architectures tested, the inference was performed independently over all data samples. To prevent overfitting on the \textit{CableInspect-AD} dataset in the zero-shot setting, we only conducted small-scale preliminary prompt optimization experiments on the MVTec AD dataset. Our experiments (results not shown) suggest that short and simple prompt instructions yield better results with the open VLMs.

The official implementation of WinCLIP is unavailable. We therefore use the implementation from \textit{anomalib} and another implementation \footnote{\url{https://github.com/caoyunkang/WinClip/blob/master/README.md}}.

\textit{Enhanced-PatchCore}, LLaVA 1.5, and BakLLaVA training and/or inference were performed on single-node NVIDIA GPUs (models A100, V100, and/or RTX8000), while CogVLM variants, inference was performed on single-node A100 GPUs (80GB). For VLMs inference, we make use of 4-bit quantization to reduce memory usage.

\begin{figure}[tb]
  \centering
  \begin{subfigure}[tb!]{\textwidth}
    \centering
    \includegraphics[width=\textwidth,trim={0 0 0 30px},clip,]
    {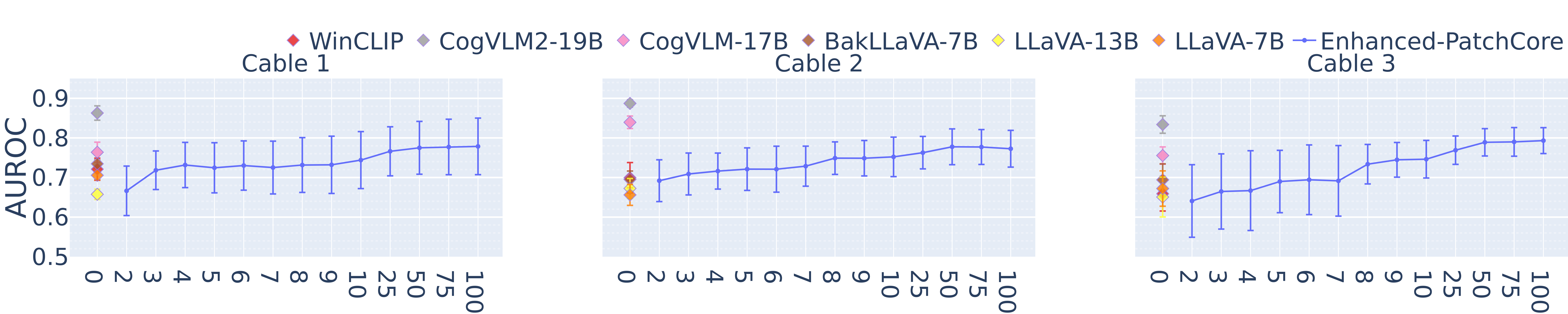}
    \caption{Mean (+/- standard deviation) AUROC over all folds vs number of train images in x-axis.}
    \label{fig:auroc}
  \end{subfigure} 
  \hspace{15px}
  \begin{subfigure}[b]{\textwidth}
    \centering
    \includegraphics[width=\textwidth,trim={0 0 0 30px},clip,]
    {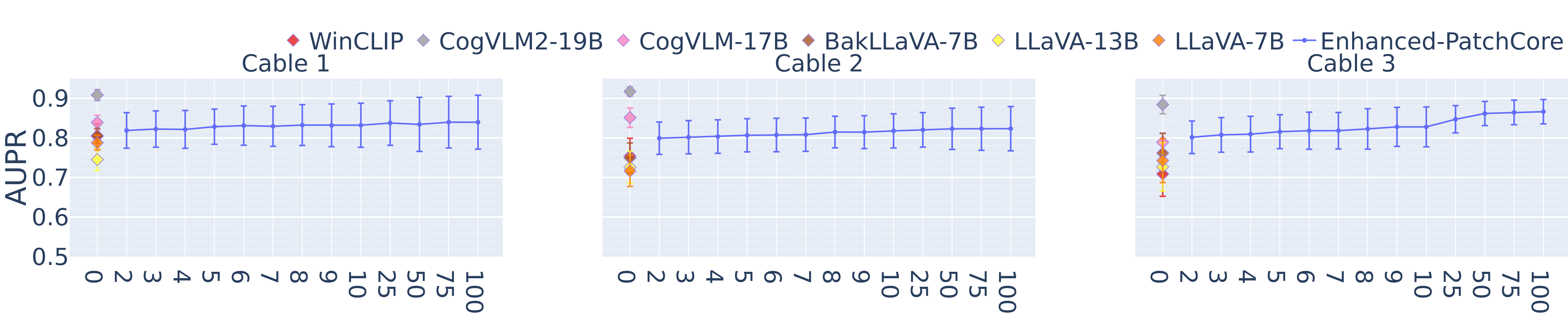}
    \caption{Mean (+/- standard deviation) AUPR over all folds vs number of train images in x-axis.}
    \label{fig:aupr}
  \end{subfigure}
  \caption{The baseline VLMs and WinCLIP in zero-shot and \textit{Enhanced-PatchCore} in few/many-shot setting results on \textit{CableInspect-AD}. (a) and (b) show  mean (+/- standard deviation) AUROC and AUPR over all folds for the three cables. The x-axis shows the number of images in the train set.
  }
  \label{fig:auroc_aupr_patchcore_raw}
\end{figure}

\begin{figure}[tb!]
  \centering
  \includegraphics[width=\textwidth,trim={0 0 0 30px},clip,]{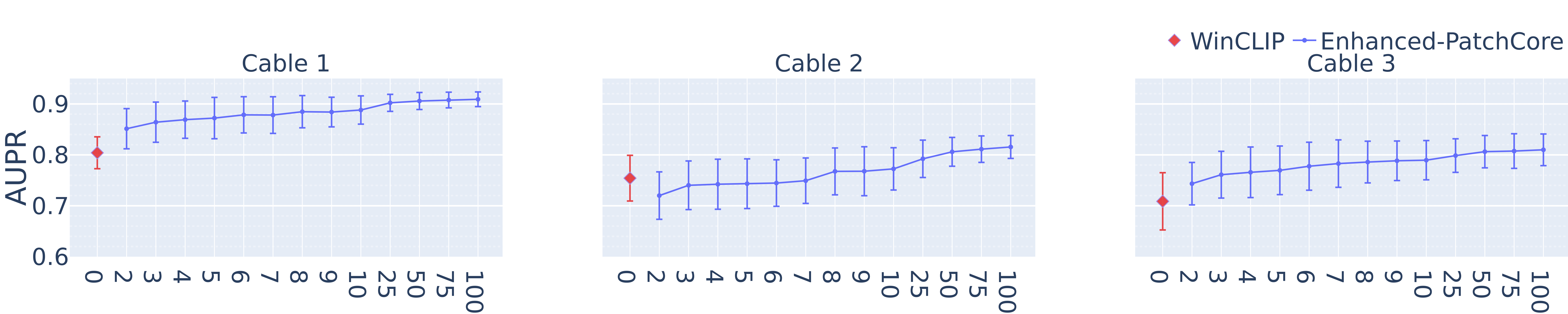}
  \caption{WinCLIP zero-shot and \textit{Enhanced-PatchCore} in few/many-shot setting results on \textit{CableInspect-AD\textunderscore cropped}. The figures show the mean (+/- standard deviation) AUPR over all folds for the three cables. The x-axis shows the number of images in the train set.
  }
  \label{fig:aupr_patchcore_crop}
\end{figure}

\subsection{Threshold-independent metrics on \textit{Enhanced-PatchCore}}
\label{sec:threshold_independent_metrics_patchcore}

\Cref{fig:auroc_aupr_patchcore_raw} shows the performance of \textit{Enhanced-PatchCore}, VLMs and WinCLIP on \textit{CableInspect-AD} using two threshold-independent metrics: Area Under the ROC curve (AUROC) and Area Under the Precision-Recall curve (AUPR). CogVLM-19B outperforms all the baseline models. However, the performance of the VLMs varies significantly. For \textit{Enhanced-PatchCore}, we see an increase in the performance i.e., the mean metric increases as the number of training images increases. However, the variance does not decrease for all cables. For example, for cables 1 and 2, the variance of AUPR increases as the number of images increases in the train set. This could be due to the variations in the background. Contrastingly, the variance decreases when the background is removed in the images, as shown in \Cref{fig:aupr_patchcore_crop}.

\subsection{Qualitative examples with VLMs}
\label{sec:qualitative_examples_vlms}

\begin{figure}[H]
\centering
\begin{tabular}{p{0.3\linewidth}|p{0.7\linewidth}} \multicolumn{2}{p{\linewidth}}{\centering\includegraphics[width=0.6\linewidth]{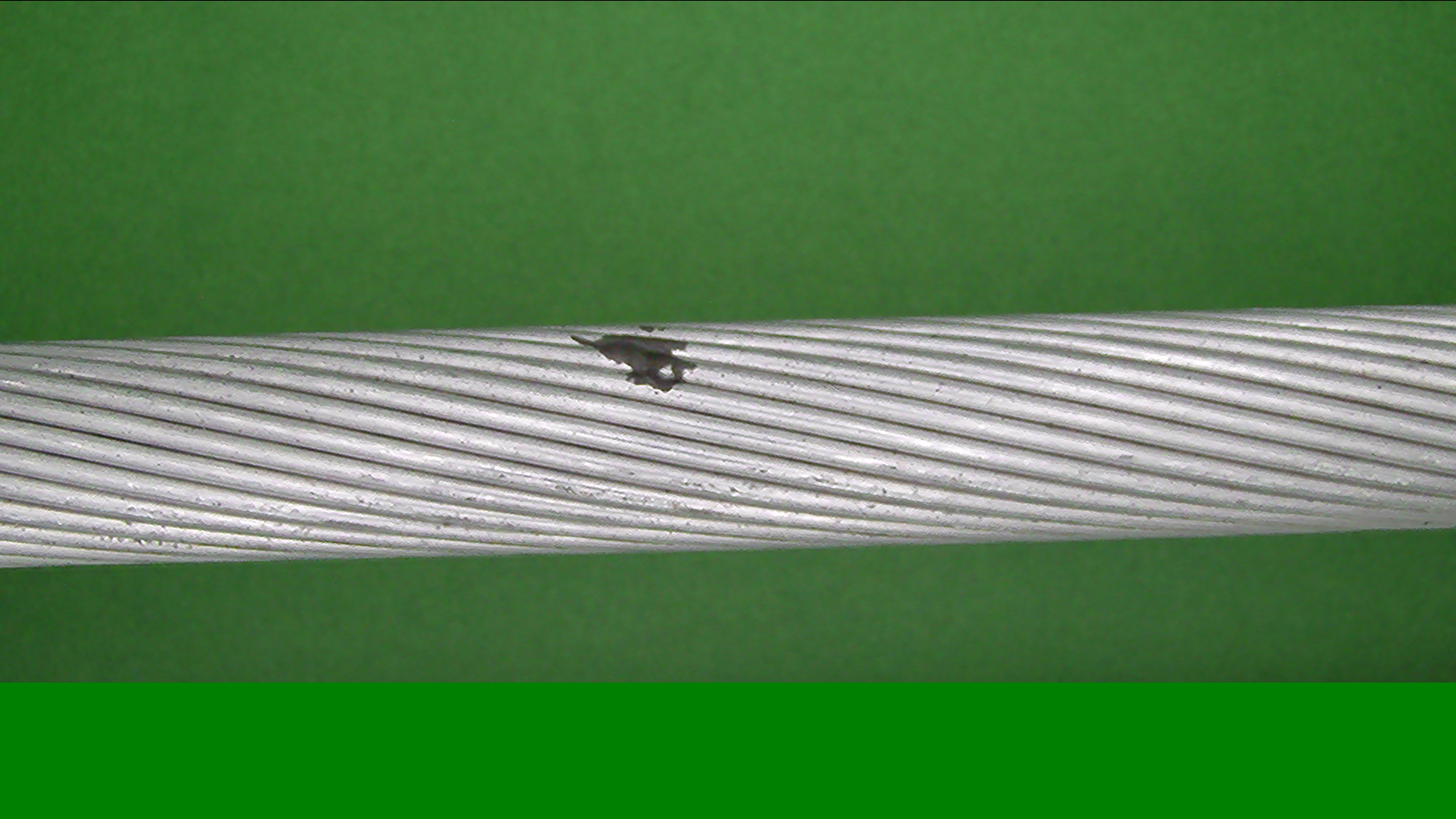}} \\ \hline
\multicolumn{2}{p{\linewidth}}{\parbox[t]{\linewidth}{\raggedright \textbf{Label:} Anomalous. Type (grade): deposit (important).}} \\ \hline
\raggedright \textbf{Prompt:} "Is there an anomaly or defect in the image? Answer by either yes or no." & \textbf{Prompt:} "Describe the content of the image." \\ \hline
\raggedright \textbf{LLaVA 1.5 7B:} \textcolor{vlm_green}{Yes.} & \textbf{LLaVA 1.5 7B:} The image features a close-up view of a
\textcolor{vlm_green}{silver wire, possibly a power cord or a cable}. There is a \textcolor{vlm_green}{small black spot on the wire}, \textcolor{vlm_orange}{which appears to be a bird dropping}. \textcolor{vlm_red}{The bird is perched on the wire, possibly taking a break or resting. The scene captures the interaction between the bird and the wire}, creating a unique and interesting
visual. \\ \hline
\raggedright \textbf{LLaVA 1.5 13B:} \textcolor{vlm_green}{Yes.} & \textbf{LLaVA 1.5 13B:} The image features a close-up view of \textcolor{vlm_green}{a silver wire}, \textcolor{vlm_orange}{possibly a speaker wire}, \textcolor{vlm_green}{with a black spot on
it}. The wire is positioned in the center of the image, and it \textcolor{vlm_green}{appears to be made of metal}. The wire is placed against a green background, which adds a contrasting color to the scene. \\ \hline
\raggedright \textbf{BakLLaVA 7B:} \textcolor{vlm_green}{Yes.} & \textbf{BakLLaVA 7B:} The image features a close-up view of a
\textcolor{vlm_green}{metal wire}, possibly a \textcolor{vlm_orange}{telephone} or \textcolor{vlm_green}{power line}, against a green background. The wire appears to be \textcolor{vlm_orange}{slightly corroded}, \textcolor{vlm_green}{with some black spots visible on its surface}. \textcolor{vlm_green}{These black spots may indicate that the wire is old or has been exposed to certain elements}, \textcolor{vlm_orange}{causing the corrosion}.
 \\ \hline
\raggedright \textbf{CogVLM 17B:} \textcolor{vlm_green}{Yes.} & \textbf{CogVLM 17B:} The image showcases a \textcolor{vlm_green}{close-up view of a metallic object, possibly a wire or cable}, \textcolor{vlm_green}{with a distinctive grooved pattern}. There's \textcolor{vlm_green}{a small, dark spot or mark on the object}. The background is a solid green color. \\ \hline
\raggedright \textbf{CogVLM2 19B:} \textcolor{vlm_green}{ Yes.} & \textbf{CogVLM2 19B:} The image showcases a \textcolor{vlm_green}{close-up view of a cylindrical object with a ridged texture}. The object appears to be \textcolor{vlm_green}{made of metal}, with \textcolor{vlm_green}{a series of parallel grooves running along its length}. There's \textcolor{vlm_green}{a small, dark, irregularly shaped mark or stain on the surface of the object}. The background is a solid green color. \\ \hline
\end{tabular}
\caption{Example of VLM outputs (sample \textit{C01\_A01\_frame\_000108}). Colors highlight the \textcolor{vlm_green}{correct}, \textcolor{vlm_orange}{incorrect but plausible}, and \textcolor{vlm_red}{incorrect} information related to the anomaly and/or cable description.}
\label{fig:vlm1}
\end{figure}

\begin{figure}[H]
\centering
\begin{tabular}{p{0.3\linewidth}|p{0.7\linewidth}} \multicolumn{2}{p{\linewidth}}{\centering\includegraphics[width=0.6\linewidth]{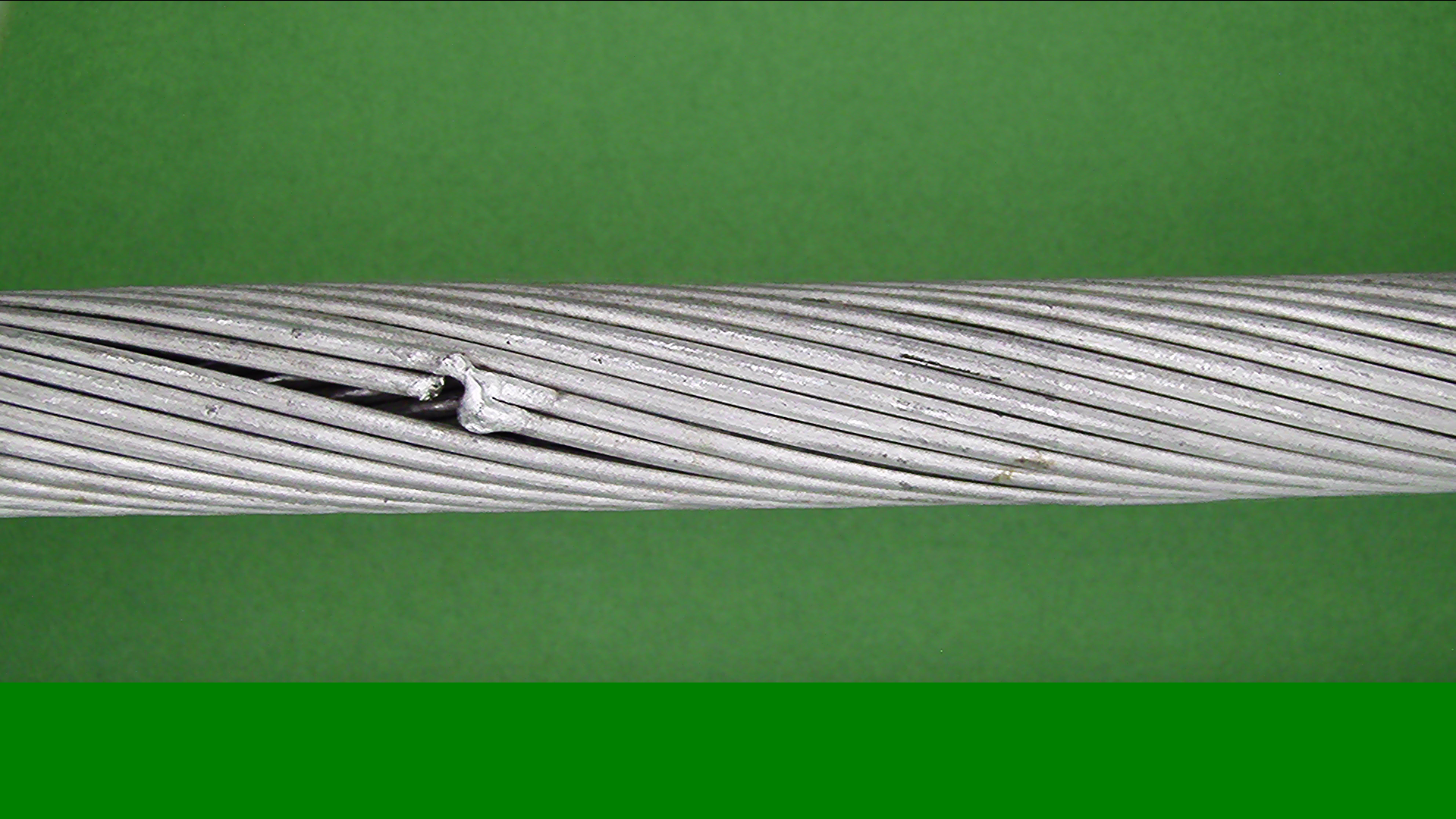}} \\ \hline
\multicolumn{2}{p{\linewidth}}{\parbox[t]{\linewidth}{\raggedright \textbf{Label:} Anomalous. Type (grade): welded strands (deep), spaced strands (light).}} \\ \hline
\raggedright \textbf{Prompt:} "Is there an anomaly or defect in the image? Answer by either yes or no." & \textbf{Prompt:} "Describe the content of the image." \\ \hline
\raggedright \textbf{LLaVA 1.5 7B:} \textcolor{vlm_green}{Yes.} & \textbf{LLaVA 1.5 7B:} The image features a close-up view of a \textcolor{vlm_green}{silver wire, possibly a power cord or a cable}. The \textcolor{vlm_green}{wire is stretched out}, and \textcolor{vlm_red}{it appears to be in good condition}. The green background provides a contrasting color to the silver wire, making it stand out. The focus of the image is on the wire, highlighting its texture and appearance. \\ \hline
\raggedright \textbf{LLaVA 1.5 13B:} \textcolor{vlm_green}{Yes.} & \textbf{LLaVA 1.5 13B:} The image features a \textcolor{vlm_green}{large, silver metal cable or wire}, \textcolor{vlm_orange}{which appears to be a piece of rope} or \textcolor{vlm_green}{a thick wire}. The cable is lying on a green background, possibly a table or a surface. The \textcolor{vlm_green}{cable is quite long, extending from the left side of the image to the right side}, covering most of the background. \\ \hline
\raggedright \textbf{BakLLaVA 7B:} \textcolor{vlm_green}{Yes.} & \textbf{BakLLaVA 7B:} The image features a close-up view of a \textcolor{vlm_green}{large, thick cable, possibly an electrical or telephone wire}, against a green background. The cable appears to be \textcolor{vlm_orange}{slightly corroded}, showing \textcolor{vlm_green}{signs of wear and tear}. \\ \hline
\raggedright \textbf{CogVLM 17B:} \textcolor{vlm_green}{Yes.} & \textbf{CogVLM 17B:} The image showcases a close-up view of a \textcolor{vlm_green}{coiled wire or cable}. The wire appears to be \textcolor{vlm_green}{made of a metallic material, possibly steel or aluminum, and is tightly wound in a helical pattern}. There's a \textcolor{vlm_green}{noticeable damage or break in one of the coils, with a small hole visible}. \\ \hline
\raggedright \textbf{CogVLM2 19B:} \textcolor{vlm_green}{Yes.} & \textbf{CogVLM2 19B:} The image showcases a close-up view of \textcolor{vlm_green}{a cylindrical object}, which appears to be \textcolor{vlm_green}{made of tightly wound metal or wire}. The object has a \textcolor{vlm_green}{shiny, metallic surface with a series of parallel ridges}. There's \textcolor{vlm_green}{a noticeable section where the winding seems to have been interrupted, revealing a gap or break in the structure}. \\ \hline
\end{tabular}
\caption{Example of VLM outputs (sample \textit{C03\_A01\_frame\_000123}). Colors highlight the \textcolor{vlm_green}{correct}, \textcolor{vlm_orange}{incorrect but plausible}, and \textcolor{vlm_red}{incorrect} information related to the anomaly and/or cable description.}
\label{fig:vlm2}
\end{figure}

\begin{figure}[H]
\centering
\begin{tabular}{p{0.3\linewidth}|p{0.7\linewidth}} \multicolumn{2}{p{\linewidth}}{\centering\includegraphics[width=0.6\linewidth]{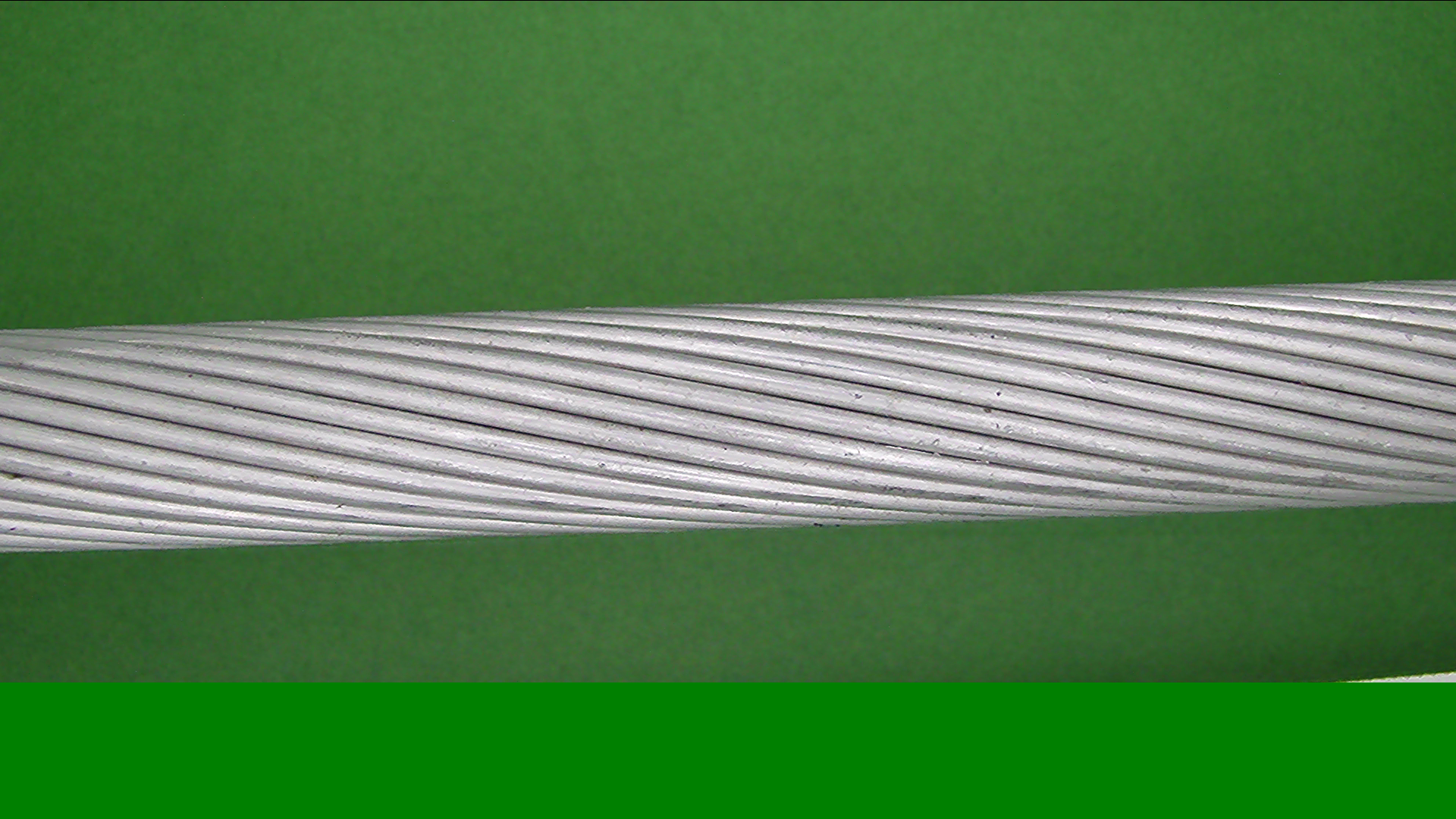}} \\ \hline
\multicolumn{2}{p{\linewidth}}{\parbox[t]{\linewidth}{\raggedright \textbf{Label:} Nominal.}} \\ \hline
\raggedright \textbf{Prompt:} "Is there an anomaly or defect in the image? Answer by either yes or no." & \textbf{Prompt:} "Describe the content of the image." \\ \hline
\raggedright \textbf{LLaVA 1.5 7B:} \textcolor{vlm_green}{No.} & \textbf{LLaVA 1.5 7B:} The image features a \textcolor{vlm_green}{large, long, and thin wire, possibly a power cable} or \textcolor{vlm_green}{a telephone wire}, \textcolor{vlm_green}{stretched across the entire frame}. The wire is \textcolor{vlm_green}{silver in color} and appears to be a \textcolor{vlm_orange}{part of a larger structure}. The green background provides a contrasting and vibrant backdrop for the wire, making it stand out prominently. \\ \hline
\raggedright \textbf{LLaVA 1.5 13B:} \textcolor{vlm_red}{Yes.} & \textbf{LLaVA 1.5 13B:} The image features a \textcolor{vlm_green}{large spool of wire, which is made of metal and has a silver color}. The spool is placed on a green background, which provides a contrasting backdrop. The wire is \textcolor{vlm_green}{wound tightly around the spool, creating a neat and organized appearance}. \\ \hline
\raggedright \textbf{BakLLaVA 7B:} \textcolor{vlm_green}{No.} & \textbf{BakLLaVA 7B:} The image features a \textcolor{vlm_green}{very long and flexible} piece of \textcolor{vlm_red}{white rope}, possibly \textcolor{vlm_red}{made of nylon or another synthetic material}. The rope is in a close-up view, making it the main focus of the image. It is set against a green background, which provides a nice contrast to the white rope. \\ \hline
\raggedright \textbf{CogVLM 17B:} \textcolor{vlm_green}{No.} & \textbf{CogVLM 17B:} The image showcases a close-up view of a \textcolor{vlm_green}{cylindrical object with a repetitive pattern of ridges or grooves}. The object appears to be \textcolor{vlm_green}{made of a metallic material, possibly steel or aluminum, given its shiny and smooth surface}. The background is a solid green color, which contrasts with the \textcolor{vlm_green}{metallic hue} of the object. \\ \hline
\raggedright \textbf{CogVLM2 19B:} \textcolor{vlm_green}{No.} & \textbf{CogVLM2 19B:} The image showcases a close-up view of \textcolor{vlm_green}{a cylindrical object with a ridged texture}. The object appears to be \textcolor{vlm_green}{made of a metallic material, with parallel ridges running along its length}. The background is a solid green color. \\ \hline
\end{tabular}
\caption{Example of VLM outputs (sample \textit{C01\_B02\_frame\_000303}). Colors highlight the \textcolor{vlm_green}{correct}, \textcolor{vlm_orange}{incorrect but plausible}, and \textcolor{vlm_red}{incorrect} information related to the anomaly and/or cable description.}
\label{fig:vlm3}
\end{figure}

\begin{figure}[H]
\centering
\begin{tabular}{p{0.3\linewidth}|p{0.7\linewidth}} \multicolumn{2}{p{\linewidth}}{\centering\includegraphics[width=0.6\linewidth]{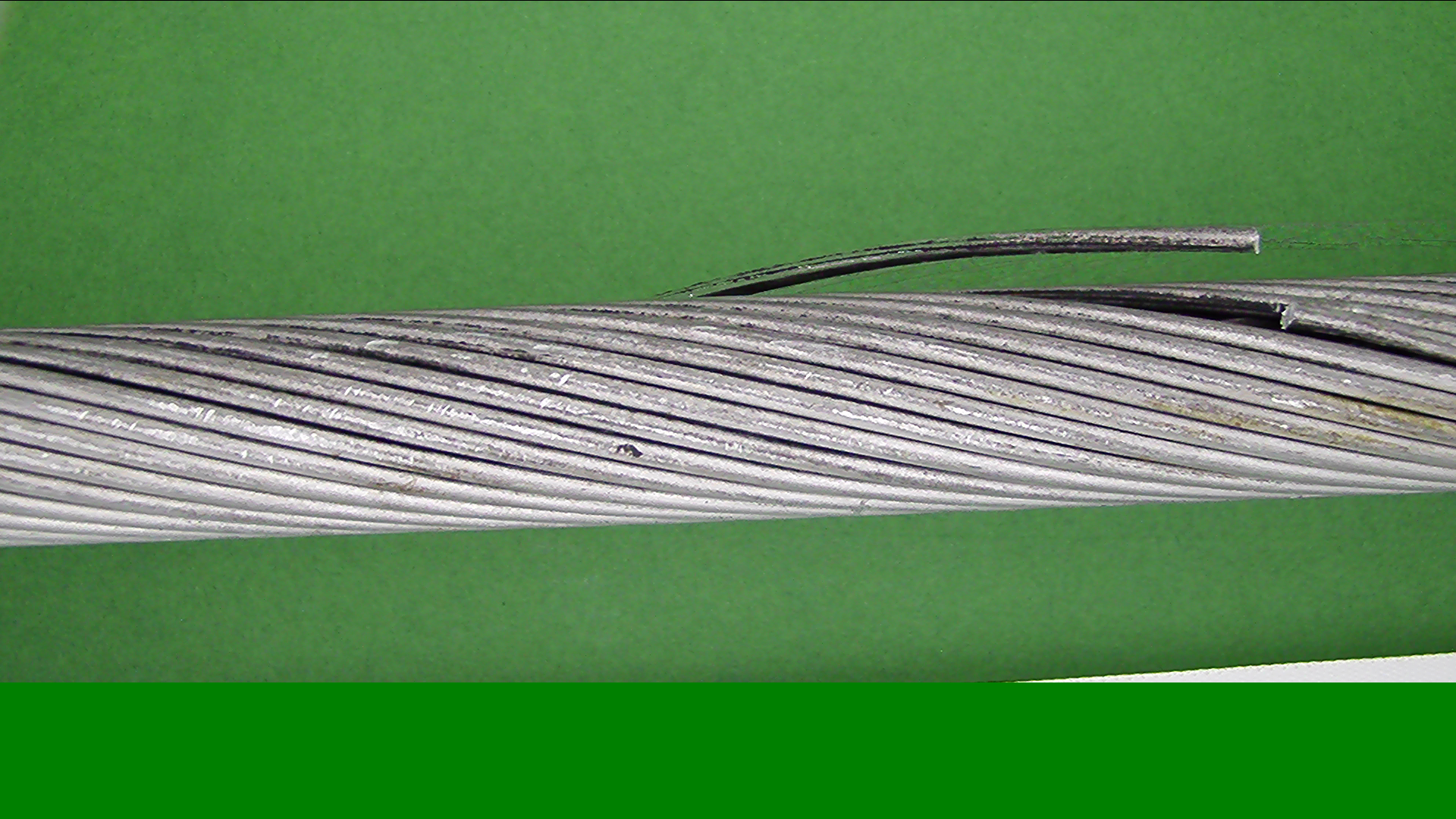}} \\ \hline
\multicolumn{2}{p{\linewidth}}{\parbox[t]{\linewidth}{\raggedright \textbf{Label:} Anomalous. Type (grade): broken strands (complete), broken strands (extracted), spaced strands (important).}} \\ \hline
\raggedright \textbf{Prompt:} "Is there an anomaly or defect in the image? Answer by either yes or no." & \textbf{Prompt:} "Describe the content of the image." \\ \hline
\raggedright \textbf{LLaVA 1.5 7B:} \textcolor{vlm_green}{Yes.} & \textbf{LLaVA 1.5 7B:} The image features a close-up view of \textcolor{vlm_green}{a wire}, possibly \textcolor{vlm_orange}{a power cord or a guitar string}. The wire is \textcolor{vlm_green}{stretched out}, and it appears to be a \textcolor{vlm_red}{black color}. The wire is the main focus of the image, and it is displayed prominently in the foreground. \\ \hline
\raggedright \textbf{LLaVA 1.5 13B:} \textcolor{vlm_green}{Yes.} & \textbf{LLaVA 1.5 13B:} The image features a close-up view of a \textcolor{vlm_green}{coiled wire}, possibly a \textcolor{vlm_orange}{speaker wire}, lying on a green background. The wire is \textcolor{vlm_green}{tightly coiled}, and its color is a \textcolor{vlm_orange}{combination of black and silver}. The wire's texture and pattern create an interesting visual effect, making it stand out against the green background. \\ \hline
\raggedright \textbf{BakLLaVA 7B:} \textcolor{vlm_green}{Yes.} & \textbf{BakLLaVA 7B:} The image features a close-up view of \textcolor{vlm_green}{a large, thick wire or cable, possibly made of metal}, lying on a green background. The wire occupies a significant portion of the frame, with \textcolor{vlm_green}{its length extending from the left to the right side of the image}. \\ \hline
\raggedright \textbf{CogVLM 17B:} \textcolor{vlm_green}{Yes.} & \textbf{CogVLM 17B:} The image showcases a close-up view of \textcolor{vlm_green}{a metallic wire or cable with a distinctive pattern of concentric circles}. The wire appears to be \textcolor{vlm_green}{slightly frayed or damaged at one end}. \\ \hline
\raggedright \textbf{CogVLM2 19B:} \textcolor{vlm_green}{Yes.} & \textbf{CogVLM2 19B:} The image showcases a close-up view of \textcolor{vlm_green}{a cylindrical object, possibly a metal rod or wire, with a series of tightly wound, parallel strands}. There's \textcolor{vlm_green}{a small, curved piece of the same material detached} and \textcolor{vlm_orange}{placed beside the main object}. The background is a solid green color. \\ \hline
\end{tabular}
\caption{Example of VLM outputs (sample \textit{C03\_A02\_frame\_000604}). Colors highlight the \textcolor{vlm_green}{correct}, \textcolor{vlm_orange}{incorrect but plausible}, and \textcolor{vlm_red}{incorrect} information related to the anomaly and/or cable description.}
\label{fig:vlm4}
\end{figure}

\Cref{fig:vlm1,fig:vlm2,fig:vlm3,fig:vlm4} showcase capabilities and limitations of VLMs on the anomaly detection task. To complement the analysis, we also provide VLMs outputs for the generic image understanding task by prompting the models to describe the content of the image. We highlight three types of output information: (i) expected/correct information that aligns with the image's content or anomaly label, (ii) incorrect but plausible information (e.g. ambiguous), and (iii) incorrect information that does not align with the image's content or anomaly label (e.g. hallucinations).

For selected examples, we can observe that CogVLM-17B and CogVLM2-19B output more precise and/or refined descriptions of the cables and their anomalies (e.g. \Cref{fig:vlm1,fig:vlm2,fig:vlm3}). In contrast, we observe that LLaVA variants generate hallucinations more often than CogVLM variants (e.g. \Cref{fig:vlm1}), and show higher inconsistency between the anomaly detection task and image description task (e.g. \Cref{fig:vlm2,fig:vlm3}).

\begin{figure}[H]
\centering
\begin{tabular}{p{0.5\linewidth}|p{0.5\linewidth}}
\raggedright \textbf{\textit{CableInspect-AD\textunderscore raw}} {\includegraphics[width=0.9\linewidth]{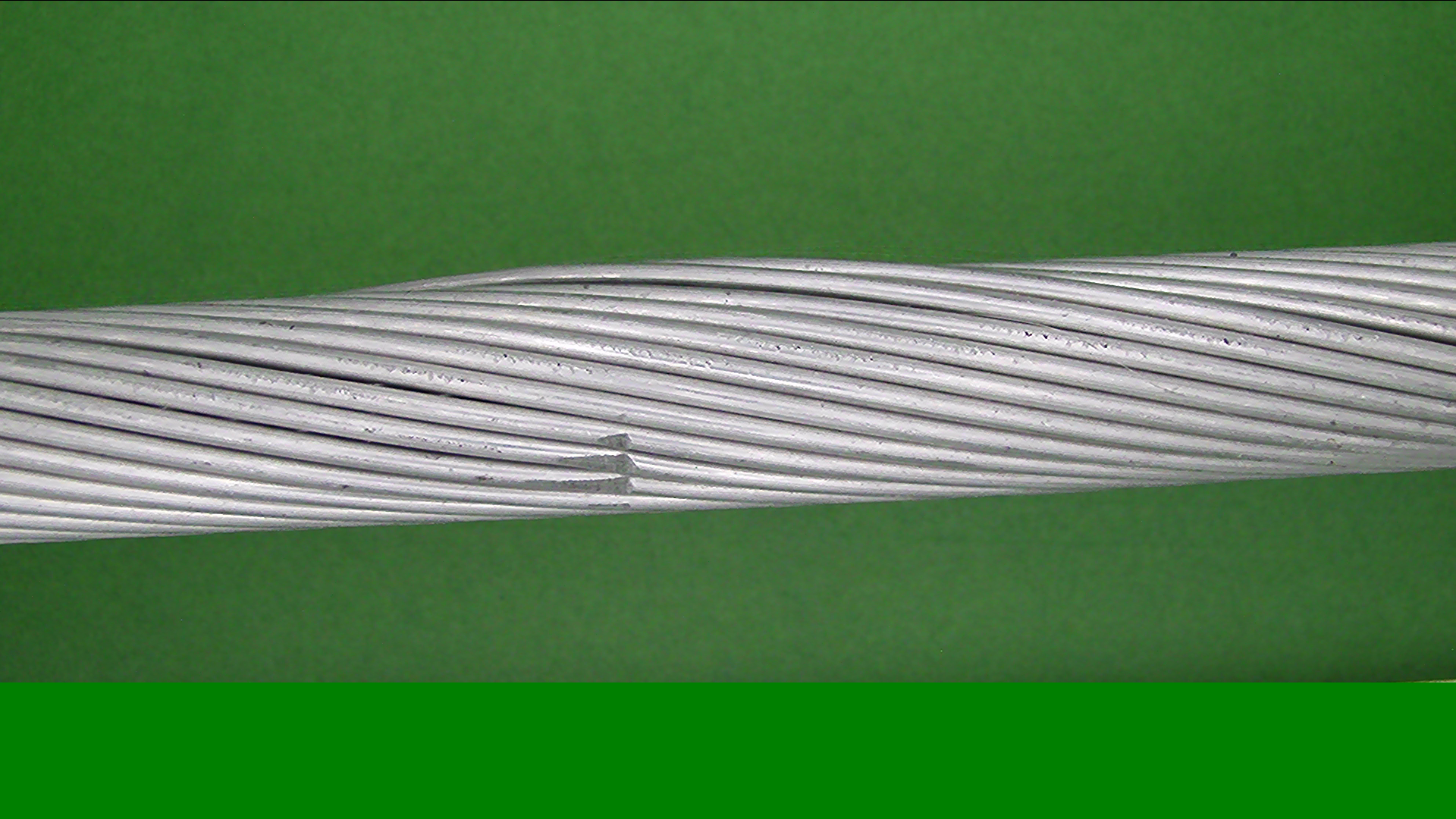}} & \textbf{\textit{CableInspect-AD\textunderscore cropped}} {\includegraphics[width=\linewidth]{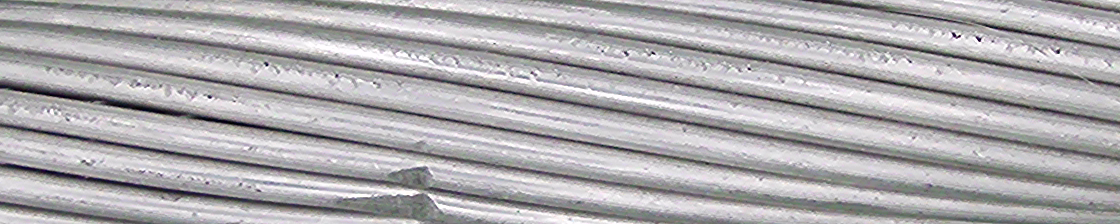}}\\ \hline
\raggedright \textbf{Label (\textit{raw}):} Anomalous. Type (grade): crushed (important), spaced strands (light), broken strands (extracted). & \textbf{Label (\textit{cropped}):} Anomalous. Type (grade): crushed (important), spaced strands (light). \\ \hline
\multicolumn{2}{p{\linewidth}}{\parbox[t]{\linewidth}{\raggedright \textbf{Prompt:} "Is there an anomaly or defect in the image? Answer by either YES or NO."}} \\ \hline
\raggedright \textbf{CogVLM 17B (\textit{raw}):} \textcolor{vlm_green}{Yes}. & \textbf{CogVLM 17B (\textit{cropped}):} \textcolor{vlm_green}{Yes}, there is a \textcolor{vlm_green}{small defect or blemish} on one of the
wires. \\ \hline
\end{tabular}
\caption{Example of CogVLM-17B outputs (sample \textit{C01\_B01\_frame\_000213}) for the \textit{raw} and \textit{cropped} versions of the sample. In this example, both versions are correctly identified as anomalous. Colors highlight the \textcolor{vlm_green}{correct}, \textcolor{vlm_orange}{incorrect but plausible}, and \textcolor{vlm_red}{incorrect} output information.}
\label{fig:cogvlm1}
\end{figure}

\begin{figure}[H]
\centering
\begin{tabular}{p{0.5\linewidth}|p{0.5\linewidth}}
\raggedright \textbf{\textit{CableInspect-AD\textunderscore raw}} {\includegraphics[width=0.9\linewidth]{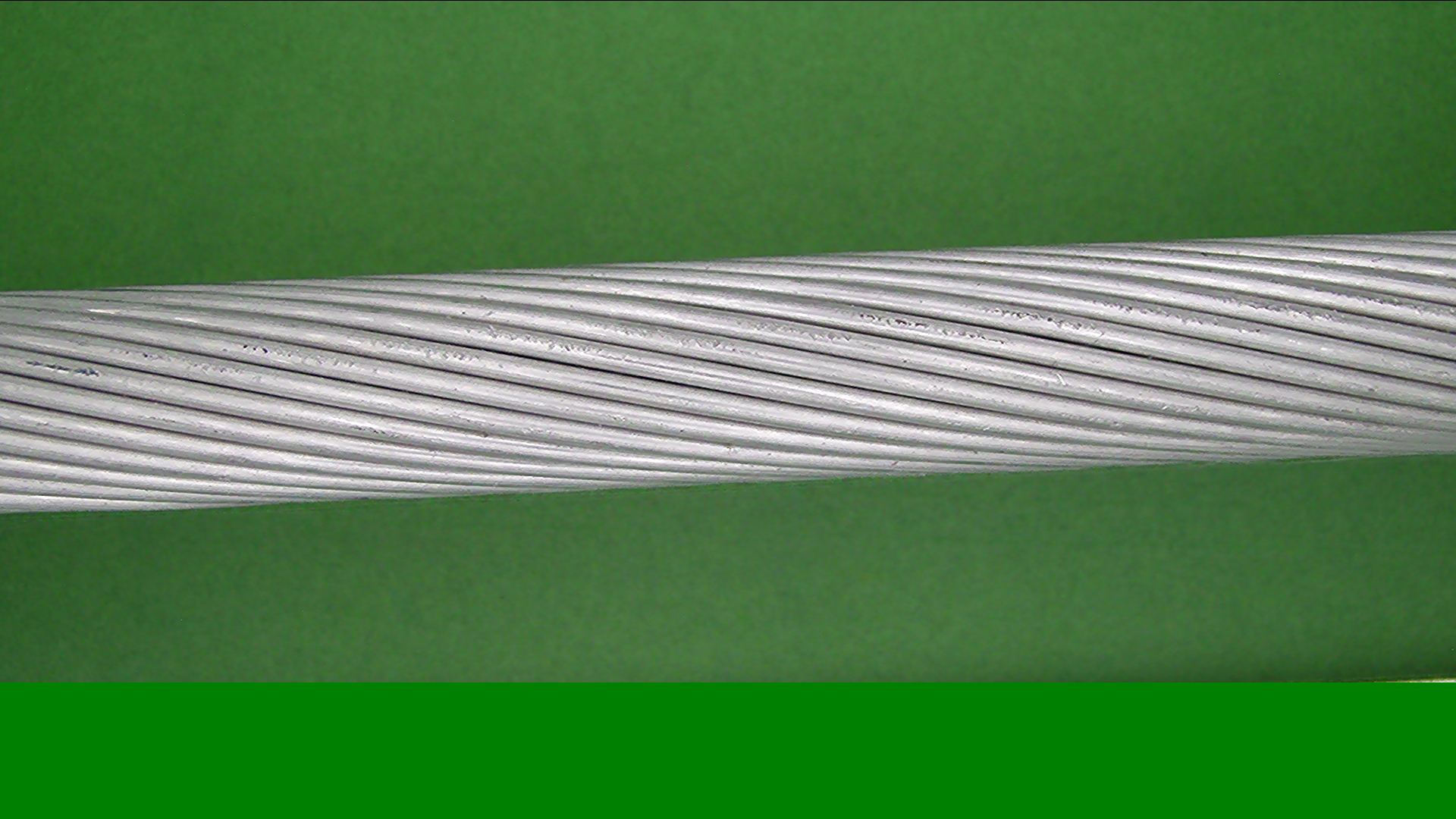}} & \textbf{\textit{CableInspect-AD\textunderscore cropped}} {\includegraphics[width=\linewidth]{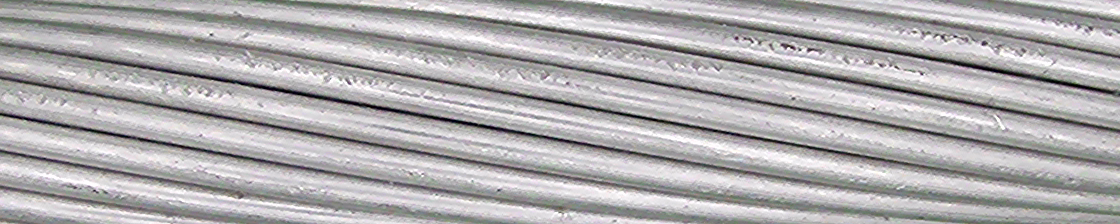}}\\ \hline
\raggedright \textbf{Label (\textit{raw}):} Nominal. & \textbf{Label (\textit{cropped}):} Nominal. \\ \hline
\multicolumn{2}{p{\linewidth}}{\parbox[t]{\linewidth}{\raggedright \textbf{Prompt:} "Is there an anomaly or defect in the image? Answer by either YES or NO."}} \\ \hline
\raggedright \textbf{CogVLM 17B (\textit{raw}):} \textcolor{vlm_green}{No}. & \textbf{CogVLM 17B (\textit{cropped}):} \textcolor{vlm_green}{No}, there doesn't appear to be any anomaly or defect in the
image. It showcases a \textcolor{vlm_green}{neatly coiled wire}.
 \\ \hline
\end{tabular}
\caption{Example of CogVLM-17B outputs (sample \textit{C01\_B01\_frame\_000246}) for the \textit{raw} and \textit{cropped} versions of the sample. In this example, both versions are correctly identified as nominal. Colors highlight the \textcolor{vlm_green}{correct}, \textcolor{vlm_orange}{incorrect but plausible}, and \textcolor{vlm_red}{incorrect} output information.}
\label{fig:cogvlm2}
\end{figure}

\begin{figure}[H]
\centering
\begin{tabular}{p{0.5\linewidth}|p{0.5\linewidth}}
\raggedright \textbf{\textit{CableInspect-AD\textunderscore raw}} {\includegraphics[width=0.9\linewidth]{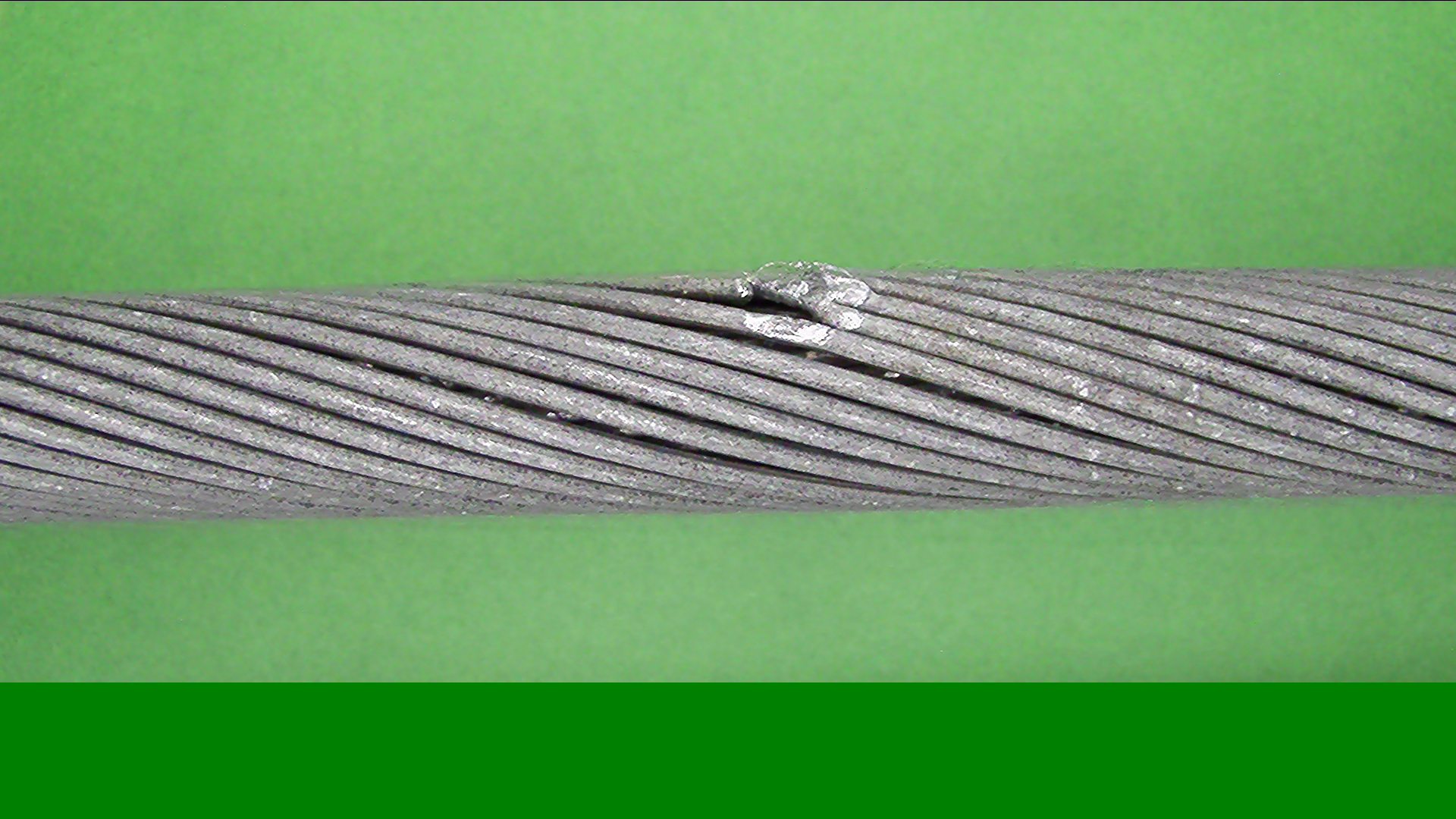}} & \textbf{\textit{CableInspect-AD\textunderscore cropped}} {\includegraphics[width=\linewidth]{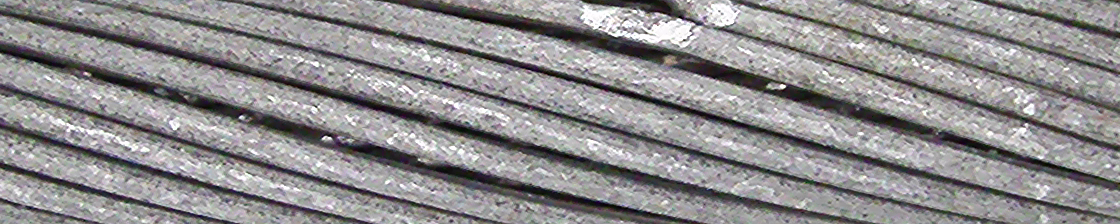}}\\ \hline
\raggedright \textbf{Label (\textit{raw}):} Anomalous. Type (grade): welded strands (deep), spaced strands (light). & \textbf{Label (\textit{cropped}):} Anomalous. Type (grade): welded strands (deep), spaced strands (light). \\ \hline
\multicolumn{2}{p{\linewidth}}{\parbox[t]{\linewidth}{\raggedright \textbf{Prompt:} "Is there an anomaly or defect in the image? Answer by either YES or NO."}} \\ \hline
\raggedright \textbf{CogVLM 17B (\textit{raw}):} \textcolor{vlm_green}{Yes}. & \textbf{CogVLM 17B (\textit{cropped}):} \textcolor{vlm_green}{Yes}, there is a \textcolor{vlm_green}{small white speck or mark} on one of the \textcolor{vlm_red}{black cords}.
 \\ \hline
\end{tabular}
\caption{Example of CogVLM-17B outputs (sample \textit{C02\_A01\_frame\_000257}) for the \textit{raw} and \textit{cropped} versions of the sample. In this example, both versions are correctly identified as anomalous. Colors highlight the \textcolor{vlm_green}{correct}, \textcolor{vlm_orange}{incorrect but plausible}, and \textcolor{vlm_red}{incorrect} output information.}
\label{fig:cogvlm3}
\end{figure}

\begin{figure}[H]
\centering
\begin{tabular}{p{0.5\linewidth}|p{0.5\linewidth}}
\raggedright \textbf{\textit{CableInspect-AD\textunderscore raw}} {\includegraphics[width=0.9\linewidth]{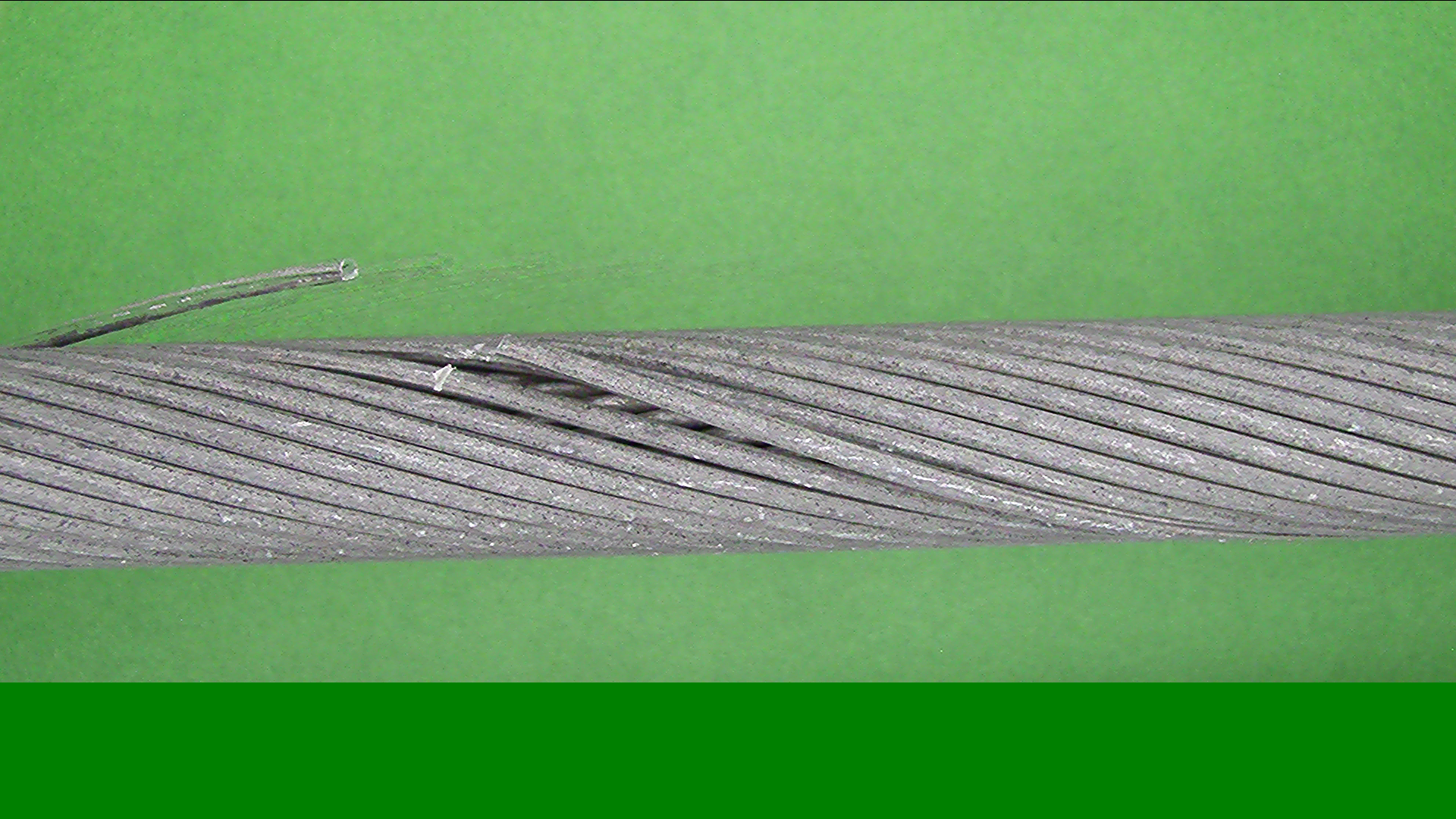}} & \textbf{\textit{CableInspect-AD\textunderscore cropped}} {\includegraphics[width=\linewidth]{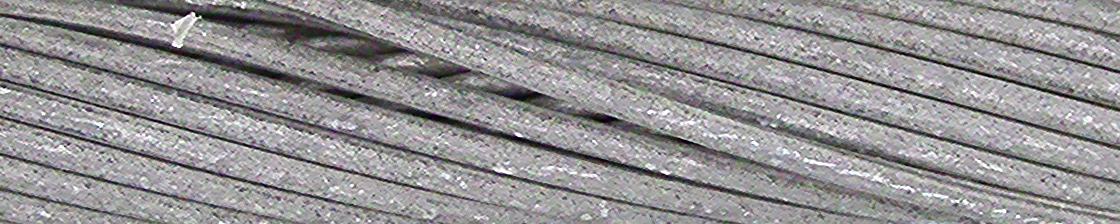}}\\ \hline
\raggedright \textbf{Label (\textit{raw}):} Anomalous. Type (grade): broken strands (complete), broken strands (extracted), spaced strands (light). & \textbf{Label (\textit{cropped}):} Anomalous. Type (grade): broken strands (complete), spaced strands (light). \\ \hline
\multicolumn{2}{p{\linewidth}}{\parbox[t]{\linewidth}{\raggedright \textbf{Prompt:} "Is there an anomaly or defect in the image? Answer by either YES or NO."}} \\ \hline
\raggedright \textbf{CogVLM 17B (\textit{raw}):} \textcolor{vlm_green}{Yes}. & \textbf{CogVLM 17B (\textit{cropped}):} \textcolor{vlm_green}{Yes}, there is a \textcolor{vlm_orange}{small white object, possibly a stitch or a piece of thread}, that is not part of the main fabric.
 \\ \hline
\end{tabular}
\caption{Example of CogVLM-17B outputs (sample \textit{C02\_A01\_frame\_000578}) for the \textit{raw} and \textit{cropped} versions of the sample. In this example, the broken strands (extracted) anomaly is lost in the \textit{cropped} version. Colors highlight the \textcolor{vlm_green}{correct}, \textcolor{vlm_orange}{incorrect but plausible}, and \textcolor{vlm_red}{incorrect} output information.}
\label{fig:cogvlm4}
\end{figure}

\begin{figure}[H]
\centering
\begin{tabular}{p{0.5\linewidth}|p{0.5\linewidth}}
\raggedright \textbf{\textit{CableInspect-AD\textunderscore raw}} {\includegraphics[width=0.9\linewidth]{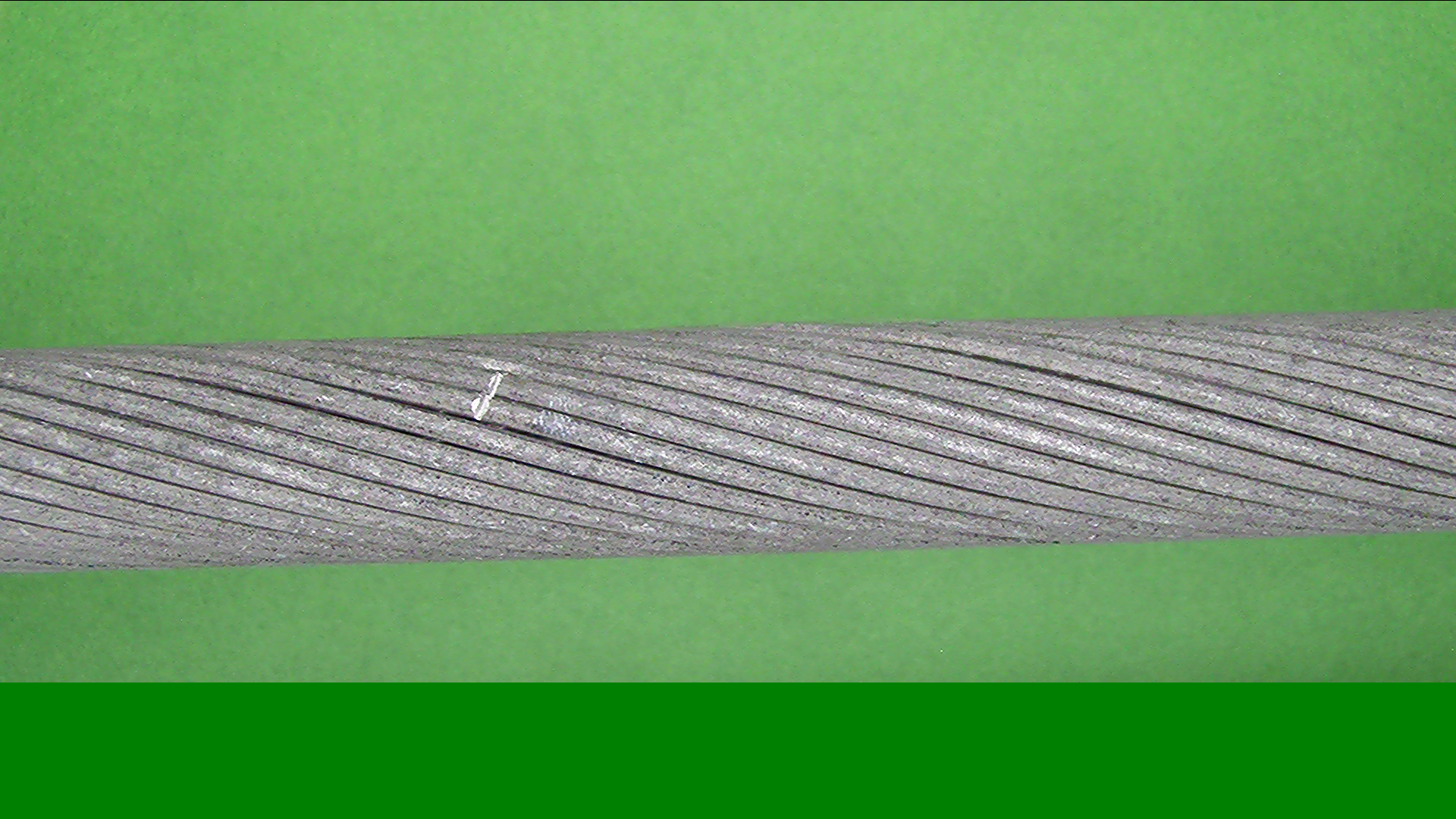}} & \textbf{\textit{CableInspect-AD\textunderscore cropped}} {\includegraphics[width=\linewidth]{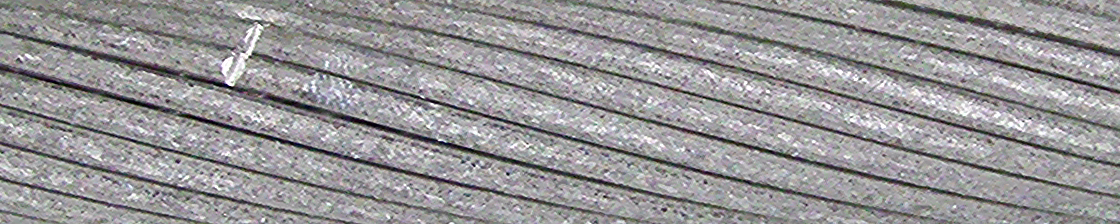}}\\ \hline
\raggedright \textbf{Label (\textit{raw}):} Anomalous. Type (grade): broken strands (partial). & \textbf{Label (\textit{cropped}):} Anomalous. Type (grade): broken strands (partial). \\ \hline
\multicolumn{2}{p{\linewidth}}{\parbox[t]{\linewidth}{\raggedright \textbf{Prompt:} "Is there an anomaly or defect in the image? Answer by either YES or NO."}} \\ \hline
\raggedright \textbf{CogVLM 17B (\textit{raw}):} \textcolor{vlm_green}{Yes}, there is an anomaly in the image. There is a \textcolor{vlm_green}{small
white mark or blemish} on the wire.
 & \textbf{CogVLM 17B (\textit{cropped}):} \textcolor{vlm_green}{Yes}, there is a \textcolor{vlm_green}{small white mark or stain} on one of the
black cables.
 \\ \hline
\end{tabular}
\caption{Example of CogVLM-17B outputs (sample \textit{C02\_A01\_frame\_000635}) for the \textit{raw} and \textit{cropped} versions of the sample. In this example, both versions are correctly identified as anomalous. Colors highlight the \textcolor{vlm_green}{correct}, \textcolor{vlm_orange}{incorrect but plausible}, and \textcolor{vlm_red}{incorrect} output information.}
\label{fig:cogvlm5}
\end{figure}

\begin{figure}[H]
\centering
\begin{tabular}{p{0.5\linewidth}|p{0.5\linewidth}}
\raggedright \textbf{\textit{CableInspect-AD\textunderscore raw}} {\includegraphics[width=0.9\linewidth]{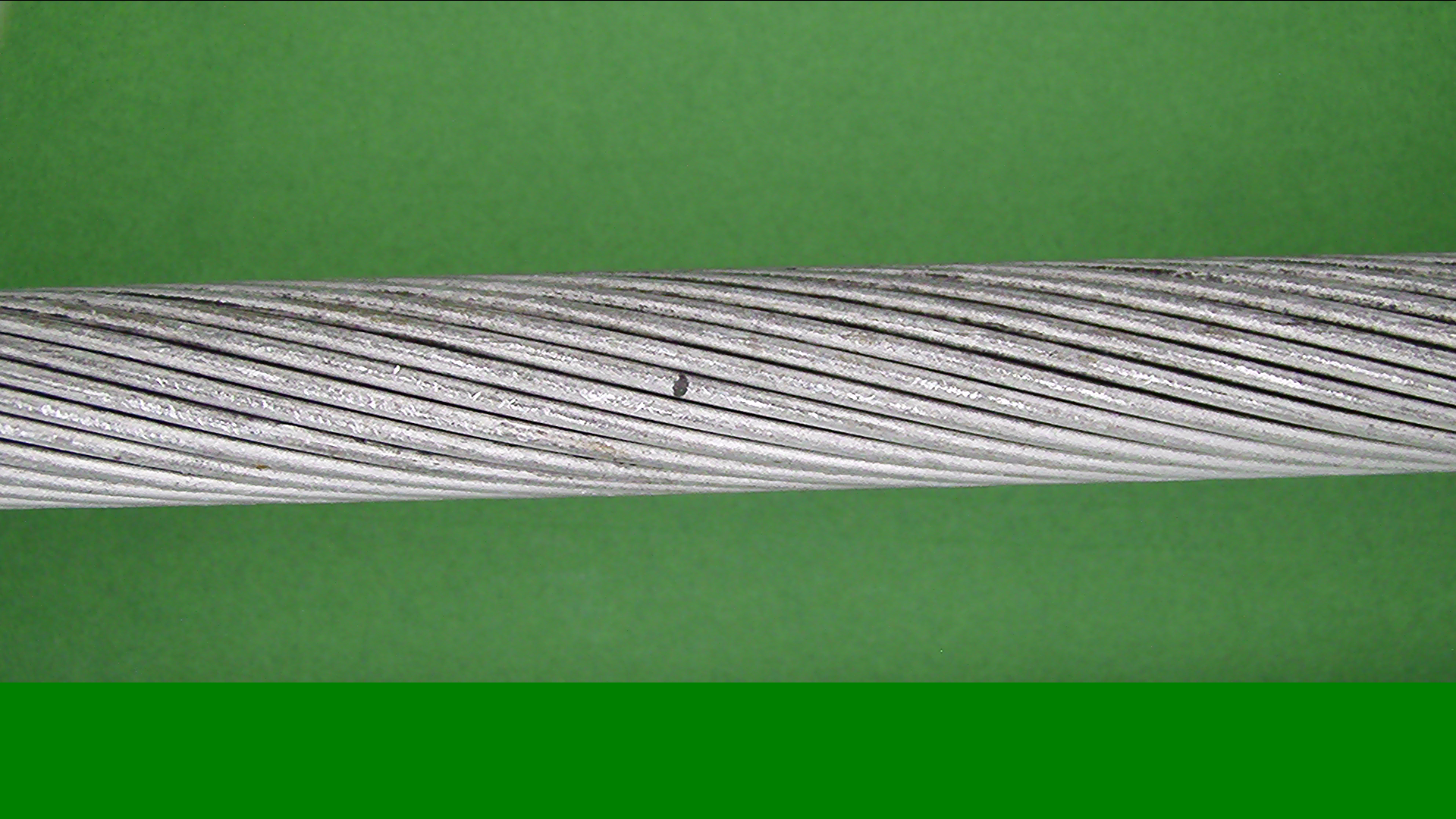}} & \textbf{\textit{CableInspect-AD\textunderscore cropped}} {\includegraphics[width=\linewidth]{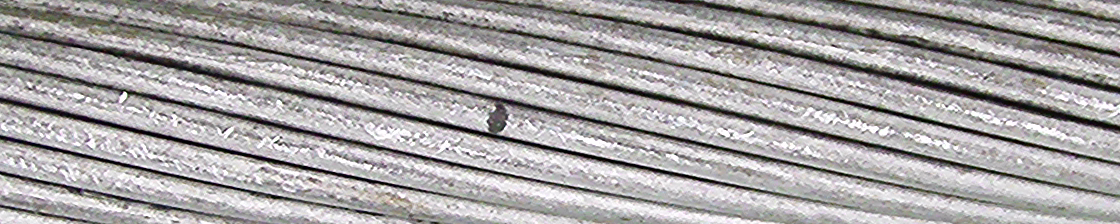}}\\ \hline
\raggedright \textbf{Label (\textit{raw}):} Nominal. & \textbf{Label (\textit{cropped}):} Nominal. \\ \hline
\multicolumn{2}{p{\linewidth}}{\parbox[t]{\linewidth}{\raggedright \textbf{Prompt:} "Is there an anomaly or defect in the image? Answer by either YES or NO."}} \\ \hline
\raggedright \textbf{CogVLM 17B (\textit{raw}):} \textcolor{vlm_red}{Yes}, there is a \textcolor{vlm_orange}{small dark spot} on the wire, which could be
a defect or an anomaly.
 & \textbf{CogVLM 17B (\textit{cropped}):} \textcolor{vlm_red}{Yes}, there is a \textcolor{vlm_orange}{small dark spot or mark} on one of the
coils.
 \\ \hline
\end{tabular}
\caption{Example of CogVLM-17B outputs (sample \textit{C03\_A01\_frame\_000429}) for the \textit{raw} and \textit{cropped} versions of the sample. In this example, both versions are identified as anomalous even though the expert labeling is nominal. Colors highlight the \textcolor{vlm_green}{correct}, \textcolor{vlm_orange}{incorrect but plausible}, and \textcolor{vlm_red}{incorrect} output information.}
\label{fig:cogvlm6}
\end{figure}

\begin{figure}[H]
\centering
\begin{tabular}{p{0.5\linewidth}|p{0.5\linewidth}}
\raggedright \textbf{\textit{CableInspect-AD\textunderscore raw}} {\includegraphics[width=0.9\linewidth]{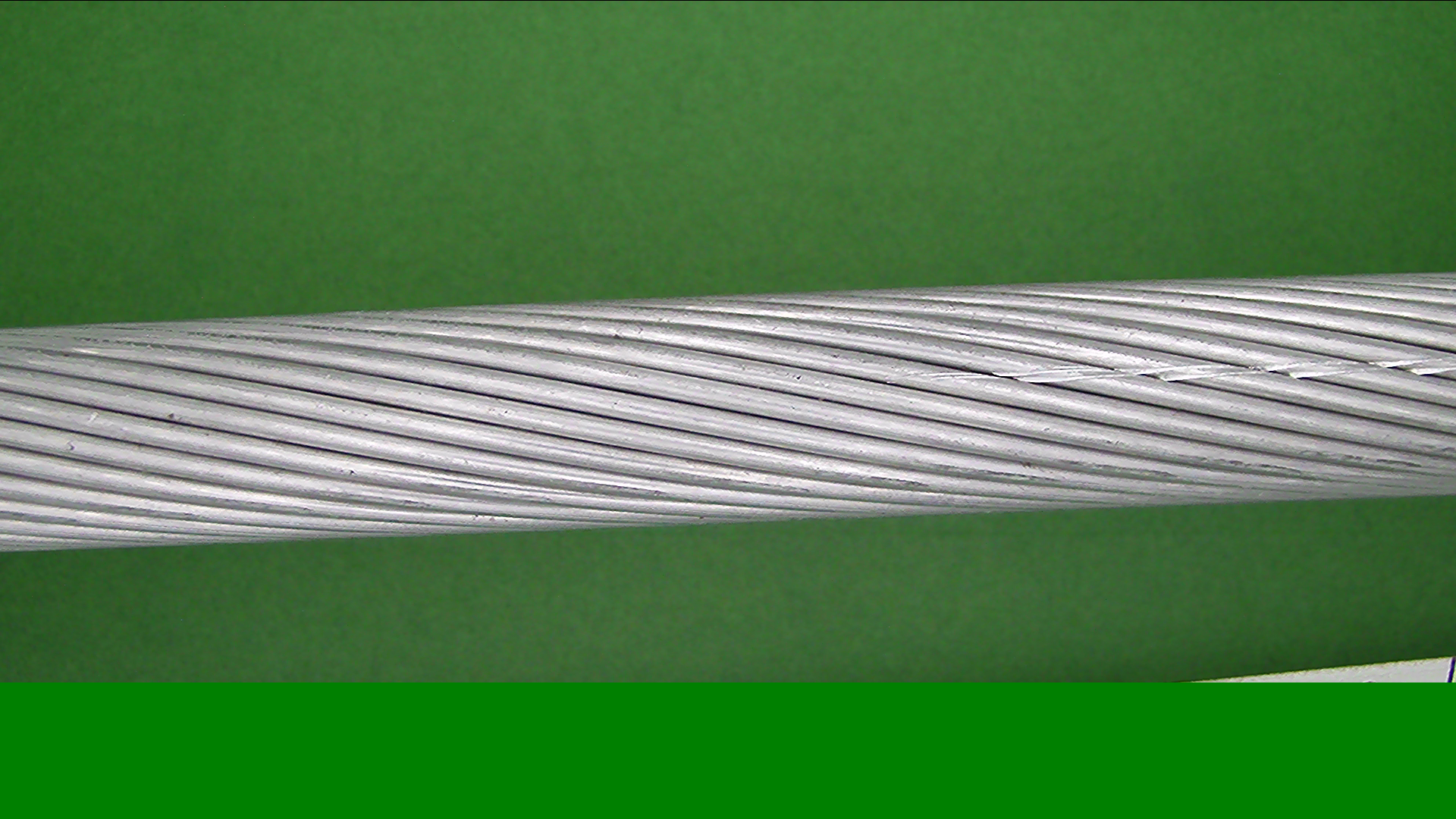}} & \textbf{\textit{CableInspect-AD\textunderscore cropped}} {\includegraphics[width=\linewidth]{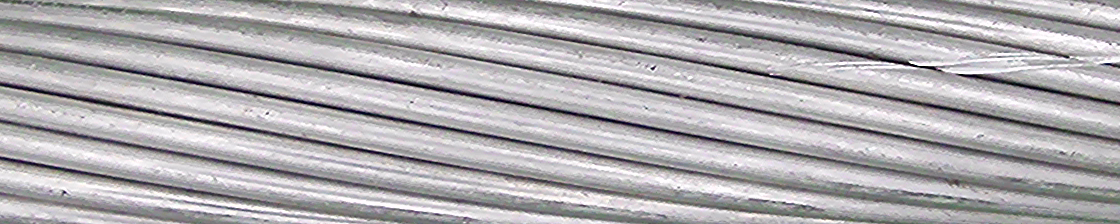}}\\ \hline
\raggedright \textbf{Label (\textit{raw}):} Anomalous. Type (grade): long scratches (light). & \textbf{Label (\textit{cropped}):} Anomalous. Type (grade): long scratches (light). \\ \hline
\multicolumn{2}{p{\linewidth}}{\parbox[t]{\linewidth}{\raggedright \textbf{Prompt:} "Is there an anomaly or defect in the image? Answer by either YES or NO."}} \\ \hline
\raggedright \textbf{CogVLM 17B (\textit{raw}):} \textcolor{vlm_green}{Yes}, there is a visible defect or anomaly in the image.
There is a \textcolor{vlm_green}{visible crack or break} in the wire, which is not
uniform and deviates from the regular pattern of the other
wires.
 & \textbf{CogVLM 17B (\textit{cropped}):} \textcolor{vlm_green}{Yes}, there is a \textcolor{vlm_green}{small white mark or defect} on one of the
coils.
 \\ \hline
\end{tabular}
\caption{Example of CogVLM-17B outputs (sample \textit{C01\_B02\_frame\_000195}) for the \textit{raw} and \textit{cropped} versions of the sample. In this example, the anomaly is correctly identified in both versions. Colors highlight the \textcolor{vlm_green}{correct}, \textcolor{vlm_orange}{incorrect but plausible}, and \textcolor{vlm_red}{incorrect} output information.}
\label{fig:cogvlm7}
\end{figure}

\begin{figure}[H]
\centering
\begin{tabular}{p{0.5\linewidth}|p{0.5\linewidth}}
\raggedright \textbf{\textit{CableInspect-AD\textunderscore raw}} {\includegraphics[width=0.9\linewidth]{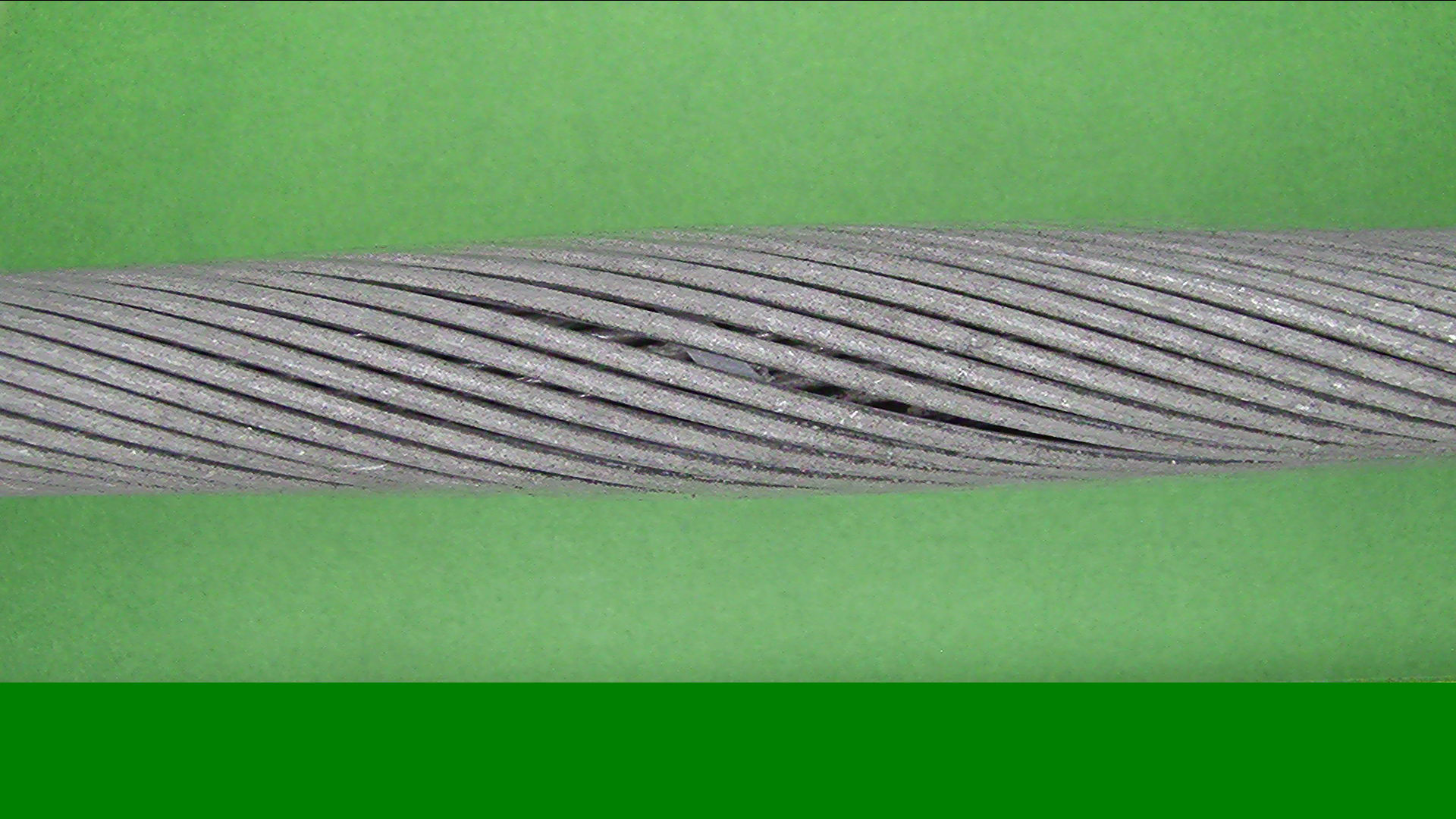}} & \textbf{\textit{CableInspect-AD\textunderscore cropped}} {\includegraphics[width=\linewidth]{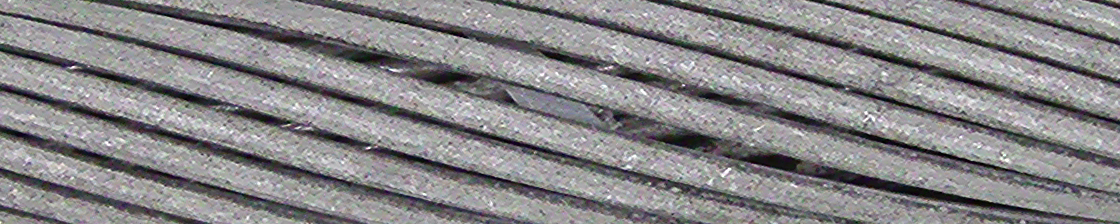}}\\ \hline
\raggedright \textbf{Label (\textit{raw}):} Anomalous. Type (grade): spaced strands (important). & \textbf{Label (\textit{cropped}):} Anomalous. Type (grade): spaced strands (important). \\ \hline
\multicolumn{2}{p{\linewidth}}{\parbox[t]{\linewidth}{\raggedright \textbf{Prompt:} "Is there an anomaly or defect in the image? Answer by either YES or NO."}} \\ \hline
\raggedright \textbf{CogVLM 17B (\textit{raw}):} \textcolor{vlm_green}{Yes}.
 & \textbf{CogVLM 17B (\textit{cropped}):} \textcolor{vlm_red}{No}, there doesn't appear to be any anomaly or defect in the
image. It showcases a \textcolor{vlm_red}{neatly coiled material}.
 \\ \hline
\end{tabular}
\caption{Example of CogVLM-17B outputs (sample \textit{C02\_B02\_frame\_000746}) for the \textit{raw} and \textit{cropped} versions of the sample. In this example, the anomaly related to the spaced strands is not detected in the \textit{cropped} version. Colors highlight the \textcolor{vlm_green}{correct}, \textcolor{vlm_orange}{incorrect but plausible}, and \textcolor{vlm_red}{incorrect} output information.}
\label{fig:cogvlm8}
\end{figure}

Similarly, \Cref{fig:cogvlm1,fig:cogvlm2,fig:cogvlm3,fig:cogvlm4,fig:cogvlm5,fig:cogvlm6,fig:cogvlm7,fig:cogvlm8} show CogVLM-17B predictions for the anomaly detection task on the same sample ID taken from the two different dataset versions, \textit{CableInspect-AD\textunderscore raw} and \textit{CableInspect-AD\textunderscore cropped}. From the selected examples, we can observe instances where the VLM correctly identifies the sample as anomalous in both dataset versions (e.g. \Cref{fig:cogvlm1,fig:cogvlm3,fig:cogvlm4,fig:cogvlm5,,fig:cogvlm7}), and instances where the VLM correctly labels both samples as nominal (e.g. \Cref{fig:cogvlm2}). We can also observe that strand-related anomalies (e.g. spaced strands, broken strands) are more likely to be ignored in the cropped version. For instance, CogVLM-17B seems to be more sensitive to discoloration (e.g. welded strands accompanied by discoloration, see \Cref{fig:cogvlm1,fig:cogvlm3,fig:cogvlm4,fig:cogvlm5}) rather than pure structural changes (e.g. spaced strands, see \Cref{fig:cogvlm8}). Finally, nominal samples that present small defects not considered anomalous by the experts can lead to false positives in both versions of the dataset (e.g. \Cref{fig:cogvlm6}).

\newpage

\newpage

\subsection{Datasheet}
\label{sec:appendix-datasheet}

\sloppy

\definecolor{darkblue}{RGB}{46,25,110}

\newcommand{\dssectionheader}[1]{%
   \noindent\framebox[\textwidth]{%
      {\fontfamily{phv}\selectfont \textbf{\textcolor{darkblue}{#1}}}
   }
}

\newcommand{\dsquestion}[1]{%
    {\noindent \fontfamily{phv}\selectfont \textcolor{darkblue}{\textbf{#1}}}
}

\newcommand{\dsquestionex}[2]{%
    {\noindent \fontfamily{phv}\selectfont \textcolor{darkblue}{\textbf{#1} #2}}
}

\newcommand{\dsanswer}[1]{%
   {\noindent #1 \medskip}
}

\dssectionheader{Motivation}

\dsquestion{For what purpose was the dataset created?}

\dsanswer{The dataset was created to enable research on industrial anomaly detection for robotic transmission cable inspection.}

\dsquestion{Who created this dataset and on
behalf of which entity?}

\dsanswer{The dataset was created by the authors Jean-Luc Bedwani and Nicolas Pouliot on behalf of IREQ - Institut de recherche d'Hydro-Qu\'ebec.}

\dsquestion{Who funded the creation of the dataset?}

\dsanswer{The funding for the creation of the dataset was provided by  and was further supported by governmental contributions from the Ministère de l'Économie, de l'Innovation et de l'Énergie (MEIE) and Innovation, Science and Economic Development Canada (ISED).
}

\dsquestion{Any other comments?}

\dsanswer{None.}

\bigskip

\dssectionheader{Composition}

\dsquestion{What do the instances that comprise the dataset represent?}

\dsanswer{The dataset comprises high-resolution (1920$\times$1080 pixels) close-up RGBA images of three real power line cables with a focus on anomaly detection (both nominal and anomalous images are included). The anomalous images showcase seven types of anomalies with different grades comprising 110 manually created and 83 pre-existing real-world anomalies from various angles, providing a realistic setting for robotic inspection.}

\dsquestion{How many instances are there in total?}

\dsanswer{The dataset contains 4,798 annotated images: 2,639 anomalous and 2,159 nominal images. Anomalous images include 193 unique anomalies, comprising 110 manually created and 83 pre-existing real-world anomalies.}

\dsquestion{Does the dataset contain all possible instances or is it a sample of instances from a larger set?}

\dsanswer{The dataset consists of images extracted from three videos recorded at a frame rate of 30 frames per second. To facilitate anomaly annotations, one frame out of three was retained, resulting in a reduced frame rate of 10 frames per second.}

\dsquestion{What data does each instance consist of?}

\dsanswer{Each instance consists of an image accompanied by labels indicating whether it is nominal or anomalous. Anomalous instances also include expert annotations, including bounding boxes, type, and grade annotations for each anomaly. Additionally, per-pixel labels for the first video of each cable are included.}

\dsquestion{Is there a label or target associated with each instance?}

\dsanswer{Yes.}

\dsquestion{Is any information missing from individual instances?}

\dsanswer{No.}

\dsquestion{Are relationships between individual instances made explicit?}

\dsanswer{Yes. The instances come from two sides (labeled A and B) of three cables (labeled C01, C02, C03). For each cable side, three videos were recorded (labeled 01, 02, 03) at a frame rate of 30 frames per second, with frames numbered sequentially. Instances are labeled as follows: \textit{\{cable number\}\_\{cable side and video number\}\_frame\_\{frame number\}} (e.g., \textit{C02\_B01\_frame\_000376}).}

\dsquestion{Are there recommended data splits?}

\dsanswer{Yes. We propose splitting the dataset into train and test sets using a k-fold cross-validation sampling strategy based on defect identifiers where each cable side is considered independently. We provide the split in the case where the training set includes 100 images.}

\dsquestion{Are there any errors, sources of noise, or redundancies in the dataset?}

\dsanswer{Yes. The dataset contains redundancies as each cable side has been recorded three times, resulting in multiple instances covering the same part of the cable with slight variations. Also, the slow frame rate causes consecutive frames to overlap. Moreover, the dataset can contain errors and noise in the annotations, particularly for light and smaller anomalies, which can be challenging for experts to detect and annotate. The labels can be noisy, as bounding boxes lack precision, and there may be mislabeling in anomaly type/grade.}

\dsquestion{Is the dataset self-contained, or does it link to or otherwise rely on external resources?}

\dsanswer{The dataset is self-contained and does not rely on external resources.}

\dsquestion{Does the dataset contain data that might be considered confidential?}

\dsanswer{No.}

\dsquestion{Does the dataset contain data that, if viewed directly, might be offensive, insulting, threatening, or might otherwise cause anxiety?}

\dsanswer{No.}

\dsquestion{Any other comments?}
\dsanswer{None.
}

\bigskip

\dssectionheader{Collection Process}

\dsquestion{How was the data associated with each instance acquired?}

\dsanswer{The data associated with each instance was acquired through a meticulous manual process. Experts identified seven types of anomalies from actual cables in operation, each categorized by severity grades. These anomalies were manually created by experts on three real power line cables, each referenced with a unique identifier, and assigned to the corresponding anomaly types. To optimize the cable usage, experts have utilized both sides of the cables (up and down), referred to as sides A and B, respectively. The cables are suspended for image acquisition, and a realistic apparatus is used to capture the images to ensure a uniform background. Along each cable, a tape with markers identifies the location of different anomalies to ease the annotation process.}

\dsquestion{What mechanisms or procedures were used to collect the data?}

\dsanswer{For each cable side, three videos were recorded, captured at a frame rate of 30 frames per second, composed of RGBA images of 1920$\times$1080 pixels. In total, 18 videos were recorded by manually moving a camera along the cables at different speeds.}

\dsquestion{If the dataset is a sample from a larger set, what was the sampling strategy?}

\dsanswer{The dataset consists of images extracted from three videos recorded at a frame rate of 30 frames per second. To facilitate anomaly annotations, one frame out of three was retained, resulting in a reduced frame rate of 10 frames per second.}

\dsquestion{Who was involved in the data collection process and how were they compensated?}

\dsanswer{The authors Jean-Luc Bedwani and Nicolas Pouliot collected the data as part of their employment at their institution IREQ - Institut de recherche d'Hydro-Qu\'ebec.}

\dsquestion{Over what timeframe was the data collected?}

\dsanswer{The dataset was collected on actual cables within a few days.}

\dsquestion{Were any ethical review processes conducted?}

\dsanswer{Not applicable.}

\dsquestion{Any other comments?}

\dsanswer{None.}

\bigskip

\dssectionheader{Preprocessing/cleaning/labeling}

\dsquestion{Was any preprocessing/cleaning/labeling of the data done?}

\dsanswer{Yes.
    \begin{itemize}
        \item Along each cable, a tape with markers identifies the location of different anomalies to ease the annotation process. A green band is added to cover the tape during post-processing to prevent the model from exploiting this information.
        \item For each video, one frame out of three was retained, resulting in a reduced frame rate of 10 frames per second.
        \item Initial frames showing poor quality were excluded from the dataset.
        \item The labeling was done by the experts and consists of bounding boxes used to locate the anomalies. The anomaly type and grade are assigned based on the appearance of the anomaly in the image, which matches the description defined by the experts. An image containing at least one anomaly is considered anomalous. Pixel-level annotations are generated using SAM with expert annotated bounding boxes as inputs. This is followed by manual correction. Authors and other experts from IREQ - Institut de recherche d'Hydro-Qu\'ebec were involved in the labeling process. More specifically, the dataset was annotated by at least four IREQ experts who first developed and agreed on guidelines to establish a clear annotation framework. The dataset then underwent five iterative rounds of review and feedback, allowing the experts to reach a consensus. This process ensured that the final version is both reliable and reflective of real-world conditions. While very light anomalies, such as light deposits and scratches, might have been missed, the experts agreed these are not critical, as they would not require immediate repair in a real-world scenario and might even go undetected by experts. All mild and severe cases were thoroughly annotated. We did not quantify the annotation process’ performance, as it was conducted in a consensus-driven, iterative manner until an agreement was reached.
        \item Two versions of the labels are released: (1) the bounding boxes with expert annotations. (2)
        pixel-level annotations. The annotations are available in COCO format. 
    \end{itemize}
}

\dsquestion{Was the “raw” data saved in addition to the preprocessed/cleaned/labeled data?}

\dsanswer{Yes. The "raw" data was saved. However, we only provide the resampled dataset as raw version. This resampled dataset has a frame rate of 10 frames per second, a green band that covers the tape, and excludes the low-quality frames.}

\dsquestion{Is the software used to preprocess/clean/label the instances available?}

\dsanswer{Yes. We used the Computer Vision Annotation Tool (CVAT) and Python scripts.}

\dsquestion{Any other comments?}

\dsanswer{None.}

\bigskip

\dssectionheader{Uses}

\dsquestion{Has the dataset been used for any tasks already?}

\dsanswer{The dataset has been developed and used for the task of industrial anomaly detection and segmentation in the context of robotic power line cable inspection.}

\dsquestion{Is there a repository that links to any or all papers or systems that use the dataset?}

\dsanswer{Yes. Refer to the project website: \url{https://mila-iqia.github.io/cableinspect-ad/}.}

\dsquestion{What (other) tasks could the dataset be used for?}

\dsanswer{The dataset could potentially be used for research on other anomaly tasks such as type/grade classification, and localization since the annotations are compatible with these tasks as well.}

\dsquestion{Is there anything about the composition of the dataset or the way it was collected and preprocessed/cleaned/labeled that might impact future uses?}

\dsanswer{Yes. When building the dataset, we attempt to include a comprehensive range of real-world anomalies. However, this leads to a higher anomaly ratio than what is typically observed in real-world scenarios, where anomalies are rare. Additionally, despite our efforts to provide a rich set of diverse examples for effective model learning and evaluation, the dataset may not encompass every possible anomaly that may appear on a cable in a real-world setting.} 

\dsquestion{Are there tasks for which the dataset should not be used?}

\dsanswer{None that we are aware of.}

\dsquestion{Any other comments?}
\dsanswer{None.}

\bigskip

\dssectionheader{Distribution}

\dsquestion{Will the dataset be distributed to third parties outside of the entity on behalf of which the dataset was created?}

\dsanswer{Yes. The dataset is publicly available on the internet through the project website: \url{https://mila-iqia.github.io/cableinspect-ad/}.}

\dsquestion{How will the dataset be distributed?}

\dsanswer{The dataset is accessible through the project website: \url{https://mila-iqia.github.io/cableinspect-ad/}.}

\dsquestion{When will the dataset be distributed?}

\dsanswer{The dataset is available and is accessible through the project website: \url{https://mila-iqia.github.io/cableinspect-ad/}.}

\dsquestion{Will the dataset be distributed under a copyright or other intellectual property (IP) license, and/or under applicable terms of use (ToU)?}

\dsanswer{Yes. We release \textit{CableInspect-AD} in the public domain under CC BY-NC-SA 4.0 license. More details are on the project website.}

\dsquestion{Have any third parties imposed IP-based or other restrictions on the data associated with the instances?}

\dsanswer{None that we are aware of.}

\dsquestion{Do any export controls or other regulatory restrictions apply to the dataset or to individual instances?}

\dsanswer{None that we are aware of.}

\dsquestion{Any other comments?}

\dsanswer{None.}

\bigskip

\dssectionheader{Maintenance}

\dsquestion{Who will be supporting/hosting/maintaining the dataset?}

\dsanswer{The authors will support and maintain the dataset.}

\dsquestion{How can the owner/curator/manager of the dataset be contacted?}

\dsanswer{Contact the authors.}

\dsquestion{Is there an erratum?}

\dsanswer{No. Future updates (if any) will be specified in the project website.}

\dsquestion{Will the dataset be updated?}

\dsanswer{Currently, no updates are planned.}

\dsquestion{If the dataset relates to people, are there applicable limits on the retention of the data associated with the instances?}

\dsanswer{Not applicable.}

\dsquestion{Will older versions of the dataset continue to be supported/hosted/maintained?}

\dsanswer{Yes. In the case of updates, refer to the project website: \url{https://mila-iqia.github.io/cableinspect-ad/}.}

\dsquestion{If others want to extend/augment/build on/contribute to the dataset, is there a mechanism for them to do so?}

\dsanswer{Yes. Suggestions for the augmentation of the dataset can be made by contacting the authors.}

\dsquestion{Any other comments?}

\dsanswer{None.}

\newpage

\subsection{Checklist}

\begin{enumerate}

\item For all authors...
\begin{enumerate}
  \item Do the main claims made in the abstract and introduction accurately reflect the paper's contributions and scope?
    \answerYes{We propose a new challenging dataset for visual anomaly detection (VAD), featuring various types of anomalies ranging from subtle defect to large structural deformations, each presented with different grades.}
  \item Did you describe the limitations of your work?
    \answerYes{See section~\ref{sec:results_and_discussion}}
  \item Did you discuss any potential negative societal impacts of your work?
    \answerYes{See section \ref{sec:results_and_discussion}}
  \item Have you read the ethics review guidelines and ensured that your paper conforms to them?
    \answerYes{}
\end{enumerate}

\item If you are including theoretical results...
\begin{enumerate}
  \item Did you state the full set of assumptions of all theoretical results?
    \answerNA{}
	\item Did you include complete proofs of all theoretical results?
    \answerNA{}
\end{enumerate}

\item If you ran experiments (e.g. for benchmarks)...
\begin{enumerate}
  \item Did you include the code, data, and instructions needed to reproduce the main experimental results (either in the supplemental material or as a URL)?
    \answerYes{See project page.}
  \item Did you specify all the training details (e.g., data splits, hyperparameters, how they were chosen)?
    \answerYes{See Supplementary Material.}
	\item Did you report error bars (e.g., with respect to the random seed after running experiments multiple times)?
    \answerYes{We do with respect to k-fold. See section \ref{sec:results_and_discussion}}
	\item Did you include the total amount of compute and the type of resources used (e.g., type of GPUs, internal cluster, or cloud provider)?
    \answerYes{See Supplementary Material.}
\end{enumerate}

\item If you are using existing assets (e.g., code, data, models) or curating/releasing new assets...
\begin{enumerate}
  \item If your work uses existing assets, did you cite the creators?
    \answerYes{See implementation details in Supplementary Material.}
  \item Did you mention the license of the assets?
    \answerYes{}
  \item Did you include any new assets either in the supplemental material or as a URL?
    \answerYes{We release our \textit{CableInspect-AD} dataset, which can be found on our project page (see abstract).}
  \item Did you discuss whether and how consent was obtained from people whose data you're using/curating?
    \answerNA{}
  \item Did you discuss whether the data you are using/curating contains personally identifiable information or offensive content?
    \answerNA{}
\end{enumerate}

\item If you used crowdsourcing or conducted research with human subjects...
\begin{enumerate}
  \item Did you include the full text of instructions given to participants and screenshots, if applicable?
    \answerNA{}
  \item Did you describe any potential participant risks, with links to Institutional Review Board (IRB) approvals, if applicable?
    \answerNA{}
  \item Did you include the estimated hourly wage paid to participants and the total amount spent on participant compensation?
    \answerNA{}
\end{enumerate}

\end{enumerate}

\end{document}